\documentclass[10pt,twocolumn,letterpaper]{article}

\usepackage[pagenumbers]{wacv} % To force page numbers, e.g. for an arXiv version

\usepackage{algorithm}
\usepackage{algpseudocode}
\usepackage{multirow}
\usepackage[table]{xcolor}
\usepackage{arydshln}

\iftrue     
\newcommand{\dimpp}[1]{\textcolor{blue}{[DP: #1]}}
\else
\newcommand{\dimpp}[1]{\textcolor{blue}{\noindent}}
\fi

\newcommand{\mypar}[1]{\vspace{0mm}\noindent\textbf{#1}}

\definecolor{diff}{RGB}{0, 0, 255} 
\usepackage{pifont}

\usepackage{tikz}

\definecolor{wacvblue}{rgb}{0.21,0.49,0.74}
\usepackage[pagebackref,breaklinks,colorlinks,allcolors=wacvblue]{hyperref}

\def\wacvPaperID{2022} % *** Enter the WACV Paper ID here
\def\confName{WACV}
\def\confYear{2026}

\title{Sherlock Your Queries: Learning to Ask the Right Questions \\ for Dialogue-Based Retrieval}

\author{
  Dong Yun\thanks{These authors contributed equally.}, Marco Schouten\footnotemark[1], Dim Papadopoulos \\
  Technical University of Denmark (DTU) \\
  Anker Engelunds Vej 101, 2800 Kongens Lyngby \\
  {\tt\small \{s232293, marschou, dimp\}@dtu.dk}
}
\begin{document}
\maketitle
\begin{abstract}
User queries in information retrieval are often ambiguous, making it challenging for systems to identify a user's target from a single query. While recent dialogue-based interactive retrieval systems can clarify user intent, they are inefficient as they often lack an explicit strategy to ask the most informative questions. To address this limitation, we propose SherlockLLM, a dialogue-driven retrieval framework that learns an optimal questioning strategy via Reinforcement Learning (RL) and avoids the need for large-scale annotated dialogue data. In our framework, an agent is trained to generate a sequence of binary questions to efficiently narrow down the search space. To validate our approach, we introduce a benchmark with both structured and unstructured tasks. Experimental results show that SherlockLLM is a robust and efficient solution. On the structured tasks, its performance matches strong baselines and approaches the theoretical optimal defined by binary search. On the challenging unstructured task, our agent significantly outperforms these baselines, showcasing its ability to learn a highly effective information-seeking dialogue policy.
We will release the code and models upon acceptance.

\end{abstract}
    
\section{Introduction}
\label{sec:intro}
User queries in information retrieval tasks are often ambiguous or incomplete \cite{zamani2020analyzing}. This poses a significant challenge for traditional retrieval systems that rely on precise and comprehensive input. This is particularly evident in scenarios where users have a clear target in mind but lack the specific keywords to describe it, such as trying to find an image of a celebrity whose name they cannot recall. In such cases, a simple, one-shot query is often insufficient to guarantee the retrieval of a specific target image.

\begin{figure}[t]
\centering
\includegraphics[page=1, width=1\linewidth]{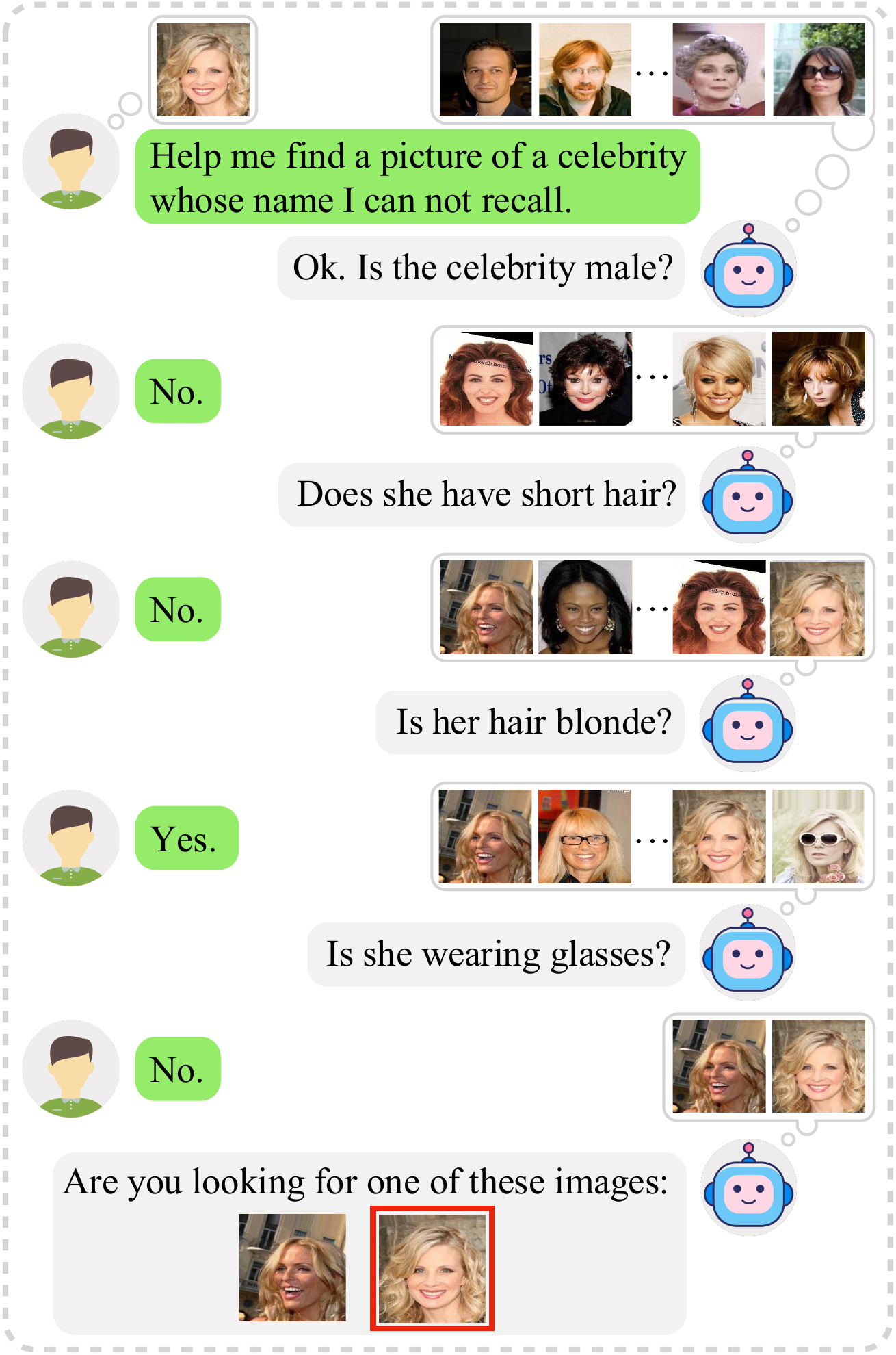}
\caption{\textbf{Our dialogue-based information retrieval agent.}
  Without precise target queries, the agent identifies the desired target through a series of yes/no questions. At each turn, the agent generates the most informative question for the user, minimizing the number of interactions needed to determine the target.}
\label{fig:sys}
\end{figure}

Dialogue-based interactive retrieval offers an effective solution by enabling systems to clarify user intent through multi-turn interactions \cite{yuan2025query,wilson-ContextualizingSearchQueries-2025}. As illustrated in \cref{fig:sys}, an agent can engage in a conversation, asking targeted questions to iteratively narrow down a large pool of candidates and pinpoint the correct item. A well-formulated question has been shown to improve retrieval efficiency by as much as 100\%~\cite{aliannejadi2019asking}. Recent approaches have used Large Language Models (LLMs) to generate clarification questions in retrieval systems \cite{siro2024agent, levy2023chatting, lee-etal-2024-interactive, li2024mediq}. However, the reasoning capabilities of LLMs do not inherently translate into effective information-seeking strategies. This is because the models lack an explicit learning signal aligned with the objective of uncertainty reduction~\cite{hu_uncertainty_2024}. Therefore, the central challenge is to generate the most informative question to minimize the number of iterations required at each dialogue turn. 

In this work, we propose \textit{SherlockLLM}, a dialogue-based retrieval framework that learns this strategic questioning ability. We formulate the task as a sequential decision-making problem solved via RL. The objective is to learn a policy that generates a sequence of questions to minimize the number of turns needed to achieve a satisfactory retrieval outcome. We train an agent to interact with a retrieval environment, where it receives rewards based on how effectively its generated questions reduce the candidate space. Through this process, the agent autonomously learns a dialogue policy optimized for retrieval efficiency.

To validate SherlockLLM's effectiveness and generalizability, we evaluate it on two distinct retrieval tasks with both structured and unstructured data. For structured data, we introduce a benchmark composed of two tasks: Guess Number and Guess Who. These tasks provide a controlled environment for evaluating the agent's ability to filter and retrieve targets via dialogue. The agent is guided by a reward function based on Expected Information Gain (EIG), and evaluates retrieval efficiency using binary search as a reference for optimal performance. For the more complex task of unstructured image retrieval, the agent's questioning policy is guided by image captions. Here, the reward function is based on the rank of the target image in a text-based search, directly optimizing for retrieval performance.

Our key contributions are threefold: 
\begin{enumerate}
\item We propose \textit{SherlockLLM}, a RL framework for dialogue-driven retrieval that autonomously learns a policy to generate optimally informative questions.
\item We introduce a benchmark comprising both structured (Guess Number and Guess Who) and unstructured (Image Retrieval) tasks to assess dialogue policy.
\item Experimental results demonstrate that our approach is a robust and generalizable solution for interactive information retrieval. On tabular tasks, its performance matches strong baselines like DeepSeek-V3.1~\cite{liu2024deepseek} and approaches the theoretical optimum. On the more challenging image retrieval task, it comprehensively outperforms strong LLMs with 96$\times$ more parameters.
\end{enumerate}

\section{Related Work} 
\mypar{Information retrieval (IR)} aims to efficiently retrieve relevant information from user queries, generally expressed in text form~\cite{datta-ImageRetrievalIdeas-2008,salton-IntroductionModernInformation-1986,salton1975vector,robertson2009probabilistic,karpukhin2020dense}.
With the advent of LLMs, IR pipelines are composed of four components: Query Rewriter, Retriever, Reranker, and Reader~\cite{zhu-LargeLanguageModels-2025}. The Query Rewriter refines the initial user query to better capture information needs. Retrievers project queries into high-dimensional vector spaces and compute relevance via inner product scoring. Rerankers apply matching methods and domain-specific objectives to refine candidate sets to improve result quality. Optionally, Readers process with LLMs retrieved information as an answer in natural language. LLMs can support different stages of the pipeline to better capture user intents and behavior~\cite{ai-InformationRetrievalMeets-2023}. 
We propose a dialogue-based framework for the \textit{Query Rewriter} that leverages LLMs to generate context-aware questions. These questions are conditioned on the history of previous user answers and guide users through a dialogue to iteratively refine and improve search results.

\mypar{LLMs as Query Rewriters.}
Query retrieval often suffers from ambiguity or incompleteness in user queries. Two common strategies to address this are query expansion and conversational query rewriting.
\textit{Query expansion}, enriches the original query with semantically related terms \cite{wang2023query2doc,jagerman2023query}.
\textit{Conversational Query Rewriters} addresses user query ambiguity through iterative, multi-turn dialogues between the IR system and user \cite{yuan2025query,wilson-ContextualizingSearchQueries-2025}. Traditional methods struggle with diverse or long-tailed dialogue sessions \cite{dai2022dialog}, whereas LLMs’ contextual understanding allows them to manage complex interactions and generate versatile queries in formats such as questions or answer-incorporated passages \cite{mao-etal-2023-large}. Moreover, RAG-based LLMs integrate explicit retrieval modules that allow the model to access external knowledge to support query generation \cite{izacard2023atlas,guu2020retrieval,borgeaud2022improving,srinivasan2022quill,ma2023query}. 
Building on these advances, our approach employs a Query Rewriter LLM that formulates retrieval queries through dialogue by generating questions and incorporating binary (yes/no) answers from the user’s conversation history.

\mypar{Reinforcement Learning}. Query rewriters operate between users and retrieval modules, but this intermediary role often leaves them without a dedicated loss function for optimization~\cite{zhu-LargeLanguageModels-2025}. This limitation motivates the use of RL to better align query generation with downstream retrieval objectives. RL incorporates feedback from multiple sources. Ranking models provide preference signals through good-bad pairs \cite{mao2024rafe,rafailov2023direct}. Offline scoring mechanisms, such as BEQUE \cite{peng2024large}, estimate query utility based on retrieved items. Retrieval-metric rewards, inspired by DeepSeek-R1 \cite{guo2025deepseek} and adopted in DeepRetrieval \cite{jiang2025deepretrieval}, enable optimization even without supervised data. Collectively, these strategies allow query rewriters to learn policies that more effectively improve retrieval.

\mypar{Unlike prior work}, we utilize an LLM-based Query Rewriter specifically designed as a \textit{Questioner}, which we fine-tune using reinforcement learning from human feedback. Consequently, the Query Rewriter learns to generate the most relevant questions based on the conversation history and prompts the user with specific keyword attributes that are essential for successful retrieval. Overall, this framework enables the user to have only a vague idea of the query, while the LLM facilitates the interaction with the retriever to efficiently identify the target.
\section{Method}
\label{sec:method}

\begin{figure}[t]
  \centering
   \includegraphics[page=2, width=1\linewidth]{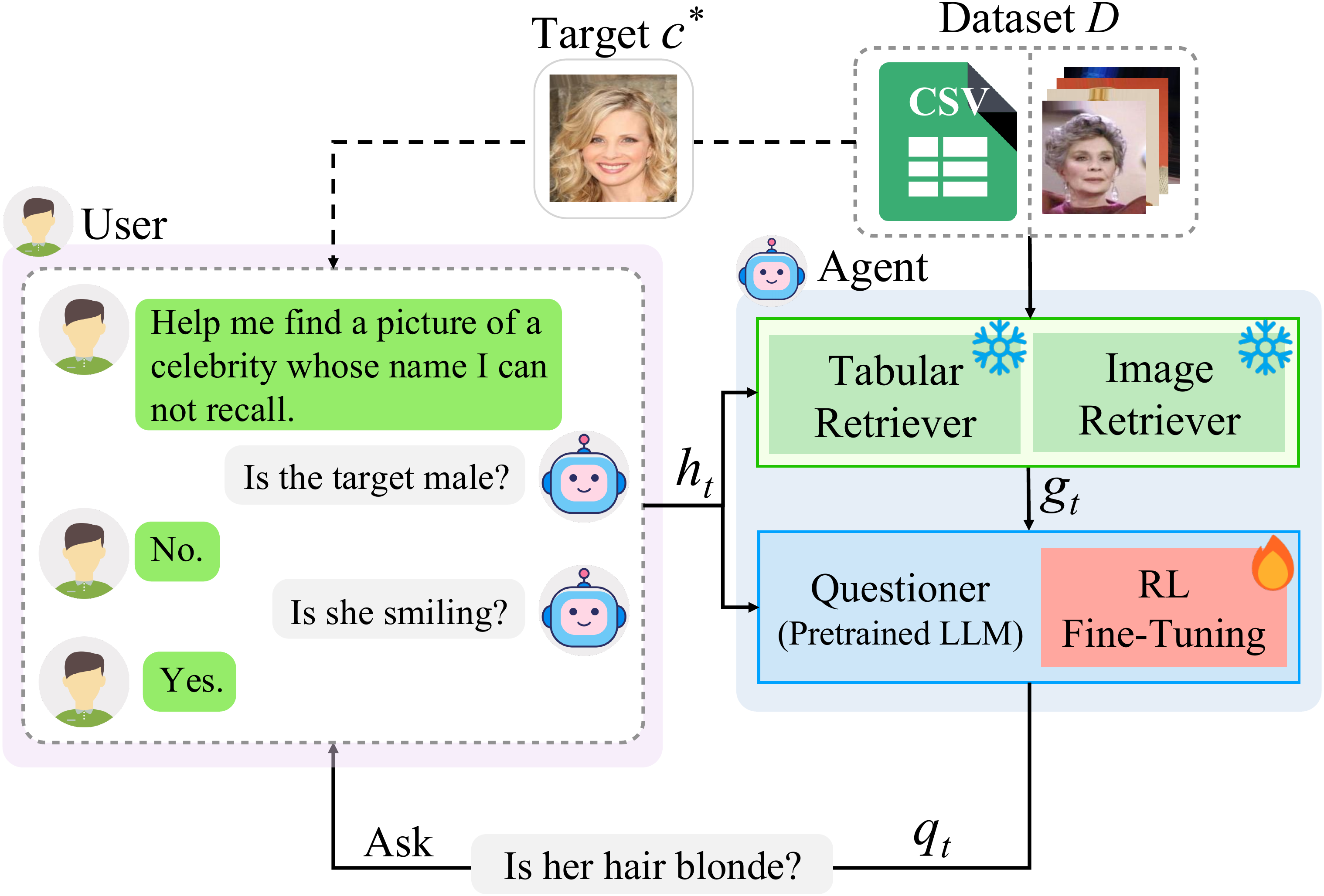}
   \caption{The overview of the pipeline. An LLM fine-tuned using reinforcement learning acts as a Questioner to ask questions, while the User answers questions as feedback.}
   \label{fig:pipeline}
\end{figure}

SherlockLLM consists of an interactive agent designed to retrieve a target item $c^*$ from a dataset $D$ through a multi-turn dialogue with a user. The agent's architecture is composed of two primary modules: a \textbf{Retriever} and a \textbf{Questioner} (\cref{fig:pipeline}). These modules operate sequentially to iteratively refine the search space and identify the target. At the beginning of an episode, a user has a target $c^*$ without a full description. The user initiates the interaction with a free-form request. At dialogue turn $t$, the Retriever takes the current dialogue history $h_t$ as input and queries $D$ to obtain results $g_t$. Subsequently, the Questioner receives both the history $h_t$ and the results $g_t$. It then generates the next yes/no question $q_t$ conditioned on both $g_t$ and $h_t$. 

A key feature of SherlockLLM is its modularity. The learning mechanism of the Questioner is domain-agnostic, while the Retriever is domain-specific. This enables seamless adaptation to different retrieval modalities by replacing the retrieval and ranking components.

\subsection{Questioner}
\label{sec:questioner}

A common strategy in dialogue-driven retrieval is to prepare a finite question pool and select a question via heuristics or a learned network \cite{keh_asking_2024, white-etal-2021-open,hu2018playing}. While effective in narrow settings, this design constrains the agent to a fixed hypothesis space: it cannot compose novel queries tailored to the unfolding dialogue, and transfers poorly across domains.

We therefore adopt a pretrained LLM as a dynamic Questioner. By leveraging broad world knowledge and generative capacity, the LLM can formulate coherent, strategic questions on the fly. 
We model the Questioner as a policy over questions conditioned on an augmented dialogue history that fuses user feedback with retrieval signals. We define the dialogue history up to turn $t$ as a sequence of question–answer pairs:

\begin{equation}
h_t = \big\{(q_0,\; a_0), \ldots,(q_{t-1},\; a_{t-1})\big\}
\label{eq:dialogue}
\end{equation}

\noindent $q_t$ is a question produced by the agent, and $a_t$ is the user’s answer. To expose the Questioner to both the user’s response and the Retriever’s result, we define an \emph{augmented answer} $\tilde a_t$:

\begin{equation}
\tilde a_t = a_t \oplus \rho\left(g_{t}\right)
\label{eq:feedback}
\end{equation}

\noindent The function $\rho(\cdot)$ renders the retrieval output $g_t$ into text (e.g., summarizing attribute distributions from tabular data or top-$k$ retrieved images). The operator $\oplus$ denotes string concatenation.

\begin{algorithm}[t]
\caption{Keyword-Conditioned CLIP Retriever}
\label{alg:clip_retriever}
\textbf{Inputs:} $h_t$ (dialogue history), 
$I = \{I_1,\dots,I_n\}$ (images);
hyperparameters $\mu$ (threshold), $\beta$ (steepness), $d_0\!\in\!(0,1)$ (base discount) \\
\textbf{Output:} Ranked image list by score $\mathcal{S}$
\begin{algorithmic}[1]
\State $(K^+, K^-)\gets \textproc{KeywordParser}(h_t)$ \Comment{Extract positive keywords $K^+$ and negative keywords $K^-$}
\If{$|K^+| > 0$}
  \State $t^+ \gets \textproc{Join}(K^+,\text{``,''})$ \Comment{Join positive keywords to a query string}
  \State $\mathcal{S}^+ \gets CLIP(I, t^+)$ 
\Else
  \State $\mathcal{S}^+ \gets \mathbf{1}_N$ 
\EndIf
\State $\mathcal{D} \gets \mathbf{1}_N$ \Comment{per-image discount vector}
\ForAll{$k^- \in K^-$}
  \State $\mathcal{S}^- \gets CLIP(I, k^-)$ 
  \State $\psi \gets 1 - (1-d_0)\cdot \sigma\!\big(\beta\,(\mathcal{S}^--\mu)\big)$ \Comment{$\sigma(x)=\tfrac{1}{1+e^{-x}}$}
  \State $\mathcal{D} \gets \mathcal{D} \odot \psi$ \Comment{elementwise product}
\EndFor
\State $\mathcal{S} \gets \mathcal{S}^+ \odot \mathcal{D}$ \Comment{final score per image}

\State \Return $\textproc{Sort}_{\downarrow}(\mathcal{S})$ \Comment{descending list by $\mathcal{S}$}
\end{algorithmic}
\end{algorithm}

The Questioner is a policy $\pi$ that generates the next question $q_t$ given the augmented dialogue history $\tilde h_t$:

\begin{equation}
q_{t} \sim \pi\left(\cdot \mid \tilde{h}_{t}\right),
\label{eq:aug-ans}
\end{equation}

The augmented history replaces each standard answer with its augmented counterpart:

\begin{equation}
\tilde h_t \;=\; \big\{(q_0,\; \tilde a_0),\; \ldots,(q_{t-1},\; \tilde a_{t-1})\big\}.
\label{eq:aug-history}
\end{equation}

\subsection{Retriever}
\label{sec:retriever}

The retriever serves as the interface between the dialogue and the underlying dataset. Its role is to interpret $h_t$, update the agent’s belief over candidates, and return the feedback to the questioner. The mechanism is similar to Retrieval-Augmented Generation (RAG). The concrete implementation is domain-specific to handle the unique characteristics of each data modality.

\mypar{Tabular Retriever.}
For structured tabular datasets, the retriever operates  \emph{deterministic logical filtering} that contracts the hypothesis space at every turn. The agent's question $q_t$ is first parsed to extract attribute–value pair(s). For instance, from the question ``Is the target's hair blonde?" the parser yields \texttt{\{hair\_color: blonde\}}. This can be implemented using either a rule-based system or a lightweight LLM.
Given the attribute–value pair(s) and the user’s answer $a_t$, the current candidate set $\mathcal{C}_t$ is filtered. The retriever outputs a new, strictly smaller candidate set $\left|\mathcal{C}_{t+1}\right|<\left|\mathcal{C}_{t}\right|$. The distribution of $\mathcal{C}_{t+1}$ is provided as feedback to the questioner. In addition, this pruned set is then used to calculate the EIG for the reward function.

\mypar{Image Retriever.}
We formalize the image retriever as a \emph{similarity-based ranker} that integrates positive and negative attributes to handle negation robustly. The retriever projects the user’s queries and all candidate images $I = \{I_1,\; \dots,\; I_n\}$ into a shared embedding space and returns a similarity score per image. Vision–language models (VLMs) such as CLIP~\cite{radford2021learning} perform well on general retrieval tasks but often degrade on queries with negation (e.g., “a man without glasses”). To address this limitation, we propose a multi-stage retrieval process that explicitly models both positive and negative attributes.

The retriever first computes positive queries for every image. As detailed in \cref{alg:clip_retriever}, our process begins by parsing the dialogue history $h_t$ to obtain two sets of keywords: positives $K^+$ and negatives $K^-$. We concatenate positive keywords into a single text query $t^+$, and compute the vector of positive similarities $\mathcal{S}^{+} \in \mathbb{R}^{n}$, the $i$-th entry gives the CLIP similarity between
$t^+$ and image $I_i$.

The retriever then converts each negative keyword into a vector of discounts $\mathcal{D}$ that penalize images similar to that keyword. For every $k^- \in K^-$, we compute a similarity vector $\mathcal{S}^{-}(k^-) \in \mathbb{R}^{n}$ whose $i$-th entry is the CLIP similarity between negative keyword $k^-$ and $I_{i}$. We transform this vector element-wise through a sigmoid gate $\sigma(\cdot)$ to obtain a discount factor $\psi(\mathcal{S}^{-}(k^-)) \in [d_0, 1]$:

\begin{equation}
\psi(x)=1-(1-d_0) \cdot \sigma\big(\beta(x-\mu)\big)
\label{eq:gate}
\end{equation}

\noindent where $d_0 \in (0,1)$ is a base discount, $\beta > 0$ defines the steepness, and $\mu \in \mathbb{R}$ is a similarity threshold. Intuitively, if an image has high similarity to a negative keyword, its corresponding discount factor $\psi(\cdot)$ approaches $d_0$, thereby applying a bounded penalty. For images with low similarity, $\psi(\cdot)$ remains close to 1, imposing a negligible penalty. 

We aggregate negative evidence by an element-wise product over all discount vectors. Let $\odot$ denote the element-wise product. The aggregate discount vector $\mathcal{D}$ is

\begin{equation}
\mathcal{D} = \bigodot_{k^- \in K^-} \psi\left(\mathcal{S}^-(k^-)\right), \; \mathcal{D}\in[d_0,1]^n
\label{eq:discount}
\end{equation}

Finally, the initial positive score $\mathcal{S}^+$ for each image is multiplied by its aggregate discount factor $\mathcal{D}$ to compute the final score $\mathcal{S}$. The retriever outputs a ranked list of all images, sorted in descending order based on $\mathcal{S}$.

\subsection{User}
\label{sec:user}
We use an LLM to simulate a user answering questions during training. Since our agent generates questions in an open-ended manner, a simple rule-based system for the simulator would be insufficient. 
The simulator workflow is as follows: First, it receives $q_t$ and the complete attribute information of $c^*$, which is dynamically formatted into a prompt. For instance, the prompt will contain the target's data (e.g., \texttt{\{"gender": "female", "hair\_color": "blonde", ...}\}) and the specific question posed by the agent. The prompt also includes a directive instructing the LLM to act as a truthful user, constraining its response to only: ``Yes'', ``No'', or ``I can't answer'', based strictly on the provided facts. This prompt is then sent to the simulator, which generates an answer $a_t$.

\subsection{Reinforcement Learning}
\label{sec:rl}

Our goal is to develop a dialogue agent that can efficiently identify a target item from a candidate set through a sequence of clarifying questions. We formulate this task as a finite Markov Decision Process (MDP), wherein a policy learns to generate questions that maximally reduce uncertainty about a hidden target.

Let $C=\{c_1, c_2, \dots, c_n\}$ be a finite set of $N$ candidate items. At the start of each dialogue episode, a single target item $c^* \in C$ is selected, which remains unknown to the agent. The agent's goal is to identify $c^*$ by engaging in a multi-turn dialogue.
The components of our RL framework are defined as follows:

\mypar{State.} At turn $t$, the state $s_t$ is represented by the dialogue history:
\begin{equation}
  s_{t} = \tilde h_{t}
  \label{eq:state}
\end{equation}
where $\tilde h_t$ is the history as defined in \cref{eq:aug-history}.

\mypar{Action.} The action is the generation of the next question $q_{t}$. This question is sampled from the policy $\pi_\theta(q_{t}|s_t)$. In our setting, the policy $\pi_\theta$ is implemented as an LLM.

\mypar{Environment.} The environment consists of a user simulator and a retriever module, which interact with the agent to generate the dialogue trajectory. An episode terminates when the agent correctly identifies the target $c^*$ or after a maximum of $T_{max}$ turns. The environment is also responsible for storing dialogue trajectories and providing a reward (\cref{sec:reward}) for each completed trajectory. 
%The reward function is detailed in .

\mypar{Learning Algorithm.} To train the agent, we employ Group Relative Policy Optimization (GRPO) \cite{shao2024deepseekmath}. GRPO estimates the baseline using intra-group relative rewards, thereby avoiding the need for a separate value function model. This approach significantly reduces memory and computational resource consumption~\cite{shao2024deepseekmath}.
During each training iteration, we sample $i$ targets. For each target $c^*$, we execute $j$ stochastic rollouts, resulting in $i\times j$ trajectories, denoted as $\{\tau_{x,y}\}_{x=1..i,\;y=1..j}$. For each $\tau$, the environment provides a trajectory-level reward $r_\tau$. We then use these collected trajectories to estimate the policy gradient and update the policy parameters $\theta$. This iterative process refines the agent's questioning strategy, making it more efficient and strategic over time. 

\subsection{Reward Functions}
\label{sec:reward}

The reward function is a principal contributing factor affecting the performance of RL agents and is also challenging to design~\cite{10766898}. Because GRPO optimizes trajectory-level rewards, we assign each episode a single scalar reward that summarizes (i) a success bonus, (ii) the average step score, and (iii) a step-length penalty. Formally, let $T$ be the number of turns in the episode $(1\le T\le T_{\max})$ and let $\epsilon_t$ denote the step score at turn $t$. We define

\begin{equation}
\bar{\epsilon} = \frac{1}{T}\sum_{t=1}^{T} \epsilon_t ,
\label{eq:avg_step}
\end{equation}

\noindent and the trajectory reward:

\begin{equation}
\mathcal{R}(\tau) =
\begin{cases}
\kappa + \bar{\epsilon} - \operatorname{Penalty}(T), & \text{success},\\
-\kappa, & \text{otherwise},
\end{cases}
\label{eq:reward}
\end{equation}

\noindent where $\kappa>0$ is a success constant. The step-length penalty scales linearly with the episode length:

\begin{equation}
\operatorname{Penalty}(T) = \alpha \cdot \frac{T}{T_{\max}}, \qquad \alpha \ge 0,
\label{eq:step_penalty}
\end{equation}

We define the step score $\epsilon_t$ differently for tabular and image settings.

\mypar{Step Score for Tabular Data.}
The tabular setting is a structured, deterministic environment. Each candidate in the current pool is described by discrete attributes. Any attribute query deterministically partitions the current candidate set into two disjoint subsets, enabling an exact accounting of uncertainty reduction.

\begin{table}[t]
\centering
\caption{Comparison to baseline for Guess Number task and Guess Who task. Oracle is the optimal baseline.}
\label{tab:tabular_res}
\setlength{\tabcolsep}{1pt}
\begin{tabular}{lccccc}
\toprule
& & \multicolumn{2}{c}{\textbf{Guess Number}} & \multicolumn{2}{c}{\textbf{Guess Who}} \\
\cmidrule(lr){3-4} \cmidrule(lr){5-6}
\textbf{Model} & \textbf{Parameters} & \textbf{SR}$\uparrow$ & \textbf{MT}$\downarrow$ & \textbf{SR}$\uparrow$ & \textbf{MT}$\downarrow$ \\
\midrule
Oracle        & - & 1.0  & 6.64  & 1.0  & 5.22 \\
\midrule
DeepSeek-V3.1 & 671B & \textbf{1.0}  & 8.02  & \textbf{1.0}  & \textbf{6.18} \\
\midrule
% \hdashline
\multirow{4}{*}{Qwen2.5}
      & 0.5B & 0.0  & 16    & 0.0  & 16 \\
      & 1.5B & 0.07 & 15.61 & 0.0  & 16 \\
      & 7B   & 0.68 & 12.43 & 0.46 & 13.26 \\
      & 32B  & 0.92 & 9.39  & 0.56 & 11.44 \\
\midrule
Qwen2.5 + SFT  & 7B & 0.88 & 10.38  & 0.48  & 11.66 \\
\midrule
Qwen2.5 + GRPO  & 7B & 0.99 & \textbf{7.62}  & \textbf{1.0}  & 6.25 \\
\bottomrule
\end{tabular}
\end{table}

We measure uncertainty using Shannon entropy~\cite{shannon1948mathematical}. This method is a widely accepted direct reward for IR~\cite{yu-etal-2020-interactive, mazzaccara-etal-2024-learning, hu_uncertainty_2024}. Let the unknown target be a random variable $c$ supported on the current candidate set $\mathcal{C}_t=\{c_{1_t},\; \dots,\; c_{n_t}\}$ at turn $t$, with belief $p_t(x)=\Pr(c=x\mid s_t)$ given state $s_t$. The conditional entropy is
\begin{equation}
H(c\mid s_t)\;=\;-\sum_{x\in\mathcal{C}_t} p_t(x)\,\log_2 p_t(x)
\end{equation}
Under a uniform belief, $H(c\mid s_t)=\log_2 n_t$.
Let $a_t\in\{\texttt{yes},\texttt{no}\}$ the user’s answer. The EIG of question $q_t$ is:

\begin{equation}
\mathrm{EIG}_t
\;=\;
H(c\mid s_t)\;-\;\mathbb{E}_{a_t\mid s_t}\!\big[\,H(c\mid s_t,a_t)\,\big]
\end{equation}

Since the answer deterministically partitions $\mathcal{C}_t$ into $\mathcal{C}_t^{\text{yes}}$ and $\mathcal{C}_t^{\text{no}}$ with sizes
$n_t^{\text{yes}}=\lvert\mathcal{C}_t^{\text{yes}}\rvert$ and $n_t^{\text{no}}=\lvert\mathcal{C}_t^{\text{no}}\rvert$, the answer probabilities under a uniform prior are
\begin{equation}
p_t^{\text{yes}}=\frac{n_t^{\text{yes}}}{n_t},
\qquad
p_t^{\text{no}}=1-p_t^{\text{yes}}=\frac{n_t^{\text{no}}}{n_t}
\end{equation}
The resulting step score is
\begin{equation}
\epsilon_t = \mathrm{EIG}_t
=
\log_2 n_t
-
\Big(p_t^{\text{yes}}\log_2 n_t^{\text{yes}} \;+\; p_t^{\text{no}}\log_2 n_t^{\text{no}}\Big)
\end{equation}

\noindent $\epsilon_t$ is maximized for a balanced split ($p_t^{\text{yes}}\approx 0.5$), achieving at most one bit of information per yes/no question.

Additionally, when a question fails the simulator-verified legality check (i.e., it is not a well-formed, answerable English interrogative), we bypass the EIG score and instead assign a soft penalty $\epsilon_t = \hat{\epsilon}$ with $-0.5 <\hat{\epsilon}< 0$.

\begin{table*}[t]
\centering
\caption{Comparison to baseline for CelebA image retrieval task.}
\label{tab:img_res}
\begin{tabular}{l ccccccc}
\toprule
\textbf{Model} & \textbf{Parameters} & \textbf{Data Size} & \textbf{SR}/\textbf{R@5}$\uparrow$ & \textbf{MT}$\downarrow$ & \textbf{MedR}$\downarrow$ & \textbf{MR}$\downarrow$ \\
\midrule
DeepSeek-V3.1  & 671B & 100 / 500 & 0.61 / 0.36 & 12.82 / 19.96 & 5.0 / 22.5 & 14.24 / 53.79  \\
\midrule
\multirow{3}{*}{Qwen2.5}
       & 1.5B  & 100 / 500 & 0.15 / 0.08 & 17.56 / 23.52 & 45.5 / 196.0 & 44.38 / 217.1 \\
       & 7B  & 100 / 500 & 0.36 / 0.23 & 15.33 / 21.52 & 17.0 / 65.0  & 25.75 / 115.30 \\
       & 32B & 100 / 500 & 0.46 / 0.24 & 13.90 / 20.89 & 8.5 / 49.5 & 20.80 / 89.42 \\
\midrule
Qwen2.5 + GRPO  & 7B & 100 / 500 & \textbf{0.90} / \textbf{0.69} &  \textbf{6.15}  / \textbf{12.21} & \textbf{3.5} / \textbf{5.0} & \textbf{6.35} / \textbf{45.11} \\
\bottomrule
\end{tabular}
\end{table*}

\mypar{Step Score for Image Data.}
In the image setting, the semantic space is high-dimensional and unstructured, making direct entropy estimation impractical. Instead, we track the agent’s progress via the \emph{rank of the ground-truth target} under a dialogue-conditioned retriever. A successful question should move the target closer to the top of the ranked list.

Given $h_t$ at turn $t$, the retriever produces similarity scores for each candidate image and induces a total ordering $\Phi_t$ by sorting candidates in descending similarity. We focus on the rank of the true target $c^\star$ under this ordering:
\begin{equation}
\Phi_t \;\in\; \{1,2,\dots,n\}, 
\qquad
\phi_t \;=\; \mathrm{rank}_{\Phi_t}(c^\star),
\end{equation}

We measure per-turn improvement by the change in \emph{log-rank}:
\begin{equation}
\epsilon_t = \Delta_t 
= \log \phi_{t-1} - \log \phi_t
\end{equation}

\noindent where $\epsilon_t>0$, when the rank improves, $\epsilon_t=0$, when it is unchanged, and $\epsilon_t<0$, when it worsens.
Similar to the tabular setting, we assign a soft penalty as a step score when an illegal question is generated.

\section{Experiments}
\label{sec:exp}

In this section, we conduct extensive experiments on two data modalities. We first introduce the datasets, and then present the experimental results and analysis.

\subsection{Datasets}

We evaluate SherlockLLM on two data modalities across three tasks. In all settings, an episode ends when the target is identified or the turn budget is exhausted.

\mypar{Guess Number.} This is a logical deduction task on a tabular dataset. In each game, a set of 100 consecutive integers is randomly chosen from a range of $[0, 1000]$, and a single number is selected as the target.

\mypar{Guess Who.} Guess Who is a multi-attribute reasoning task on a tabular dataset. We use a dataset of 36 unique characters, each defined by 9 distinct attributes (more details in \cref{sec:guesswho_dataset}).

\mypar{CelebA Image Retrieval.} This is a semantic retrieval task on an image dataset. CelebA \cite{liu2015faceattributes} is a celebrity face attributes dataset containing 202,599 images annotated with 40 binary attributes. We remove blurred images and form two sets of 100 and 500 images to assess performance under small and larger search spaces. To create rich textual captions, we first use the binary attributes to create base captions and then rewrite them with GPT-4o~\cite{hurst2024gpt} to improve fluency while preserving the encoded attributes. An episode is counted as successful when the target image is ranked within the top 5 ($\text{rank} \le 5 $) within $T_{max}$ rounds.

\subsection{Implementation Details}

\noindent\textbf{{Model.}}
We use Qwen2.5-7B-Instruct \cite{qwen2.5} as the question policy. To accelerate training and reduce computational overhead, we use a 4-bit quantized version of the model. 

\noindent\textbf{{User Simulator.}}
We use DeepSeek-V3.1~\cite{liu2024deepseek} to serve as the user simulator, responsible for answering questions.

\noindent\textbf{{Retriever.}}
For our retriever, we set the hyperparameters in~\cref{eq:gate} to $\mu=0.15$, $\beta=20$, and $d_0=0.9$. These values were empirically determined to yield optimal performance.

\noindent\textbf{{Policy Learning.}}
We fine-tune Qwen2.5-7B-Instruct using LoRA~\cite{hu2022lora} with a rank of 16 and an alpha of 32. The model is trained with the AdamW~\cite{loshchilov2017decoupled} optimizer with a learning rate of $2e^{-6}$. For the reward function, we set $\kappa$ to 2 and the step penalty coefficient to 0.7. We generate 25 dialogue trajectories for each possible target within every task. The maximum number of turns $T_{max}$ is task-dependent: for tabular tasks, $T_{max}$ is 16. For image retrieval tasks, $T_{max}$ is set to 20 for a dataset of 100 images and 25 for 500 images. All tasks are trained for 120 steps, except for the 500-image retrieval task, which is trained for 200 steps. For each task, we select the model that achieves the highest reward.

\begin{figure}[t]
    \centering
    \begin{subfigure}[b]{1\linewidth}
        \centering
        \includegraphics[page=5, width=1\linewidth]{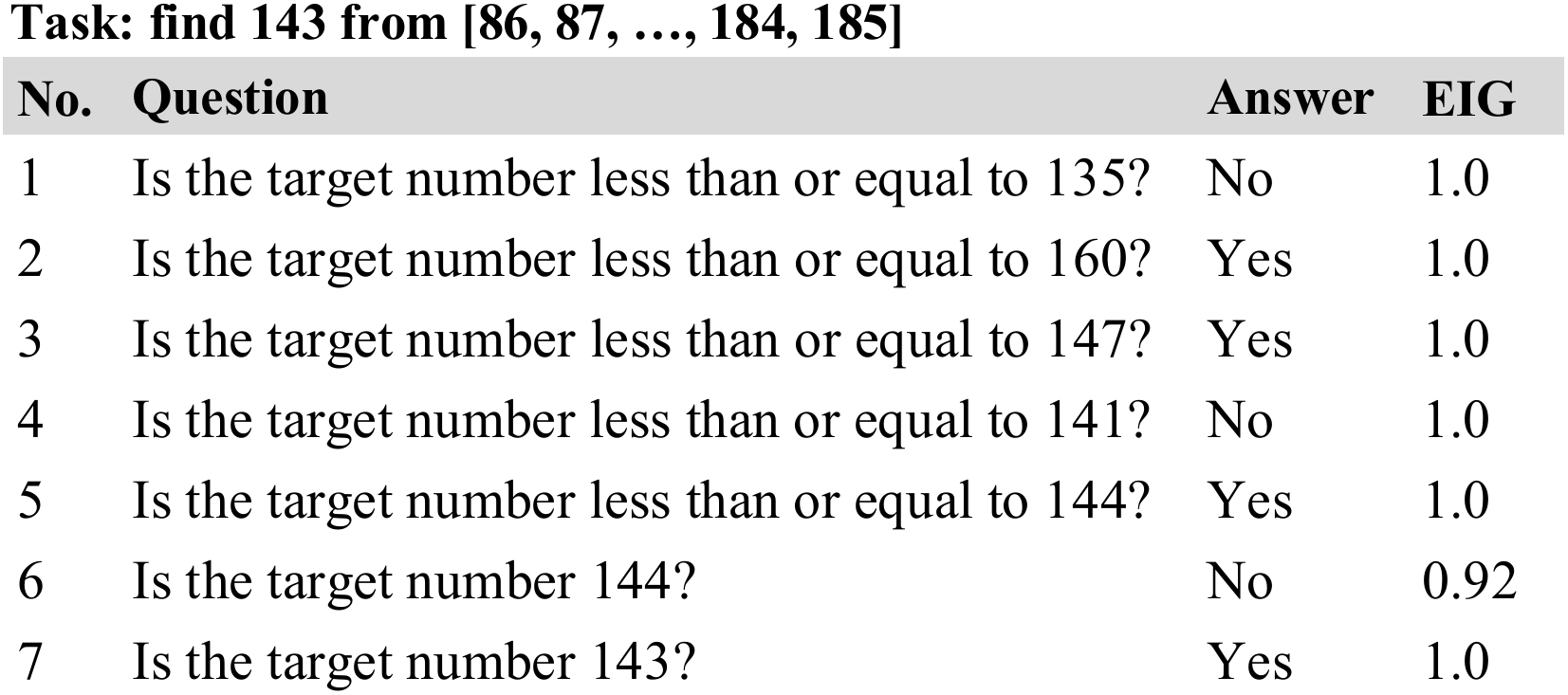}
        \caption{An example of the Guess Number task.}
        \label{fig:exp_num_sub}
    \end{subfigure}
    \par\vspace{0.5em}
    \begin{subfigure}[b]{1\linewidth}
        \centering
        \includegraphics[page=6, width=1\linewidth]{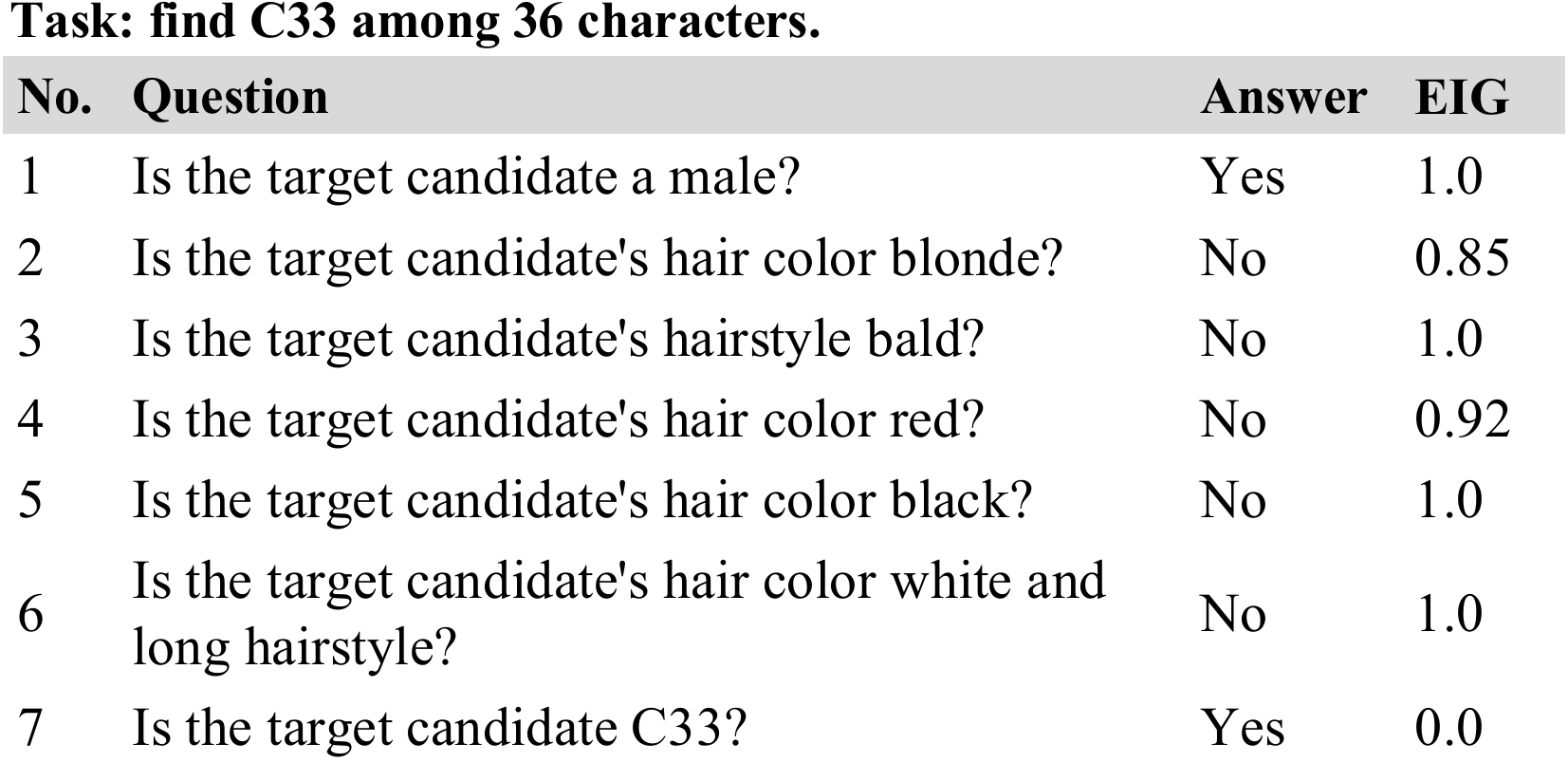}
        \caption{An example of the Guess Who task.}
        \label{fig:exp_who_sub}
    \end{subfigure}
    \caption{\textbf{Examples from the Guess Number and Guess Who tasks.} (a) illustrates the agent's strategy in Guess Number, while (b) shows its approach in Guess Who.}
    \label{fig:combined_tabular_exp}
\end{figure}

\subsection{Baselines}
\mypar{State-of-the-Art LLM (Zero-Shot).} We compare SherlockLLM against DeepSeek-V3.1 (671B parameters). This model is prompted in a zero-shot manner, using the same prompt and feedback format as our agent. For a fair comparison to other models, we do not use the ``thinking mode".

\mypar{Foundation Models (Zero-Shot).} We also compare SherlockLLM against its underlying models without any training in a zero-shot setting. We evaluate pretrained Qwen2.5-Instruct models of varying scales (0.5B, 1.5B, 7B and 32B).

\mypar{Supervised Fine-Tuning Method.} We include a baseline developed using Supervised Fine-Tuning (SFT) on tabular tasks. It directly trains a model to imitate expert behavior by using a synthetic dialogue dataset. Training details are provided in \Cref{sec:sft_guessnum} and \Cref{sec:sft_guesswho}.

\mypar{Oracle.} We define an Oracle baseline to represent the theoretical upper bound performance of our tabular tasks. The Oracle simulates a perfect decision-making process. At each round, it exhaustively searches the entire pool of single-attribute questions and selects the one that yields the maximum EIG. This question is the most informative one, and ideally, it halves the remaining candidate set. This process continues until the target's probability becomes 1. 

\subsection{Metrics}

For all tasks, we report the Success Rate (SR), which is the percentage of dialogues that successfully meet the task-specific completion criteria, and the Mean Turns (MT), which measures the mean number of turns for all dialogues.
For the CelebA Image Retrieval task, we also report common retrieval metrics: Recall@K (R@K) for K=5, representing the percentage of trials where the target is ranked within the top 5. By this definition, R@5 is equivalent to SR. Finally, we report the Median Rank (MedR) and the Mean Rank (MR) of the target image across all trials.

\begin{figure}[t]
    \centering
    \begin{subfigure}[b]{1\linewidth}
        \centering
        \includegraphics[page=3, width=1\linewidth]{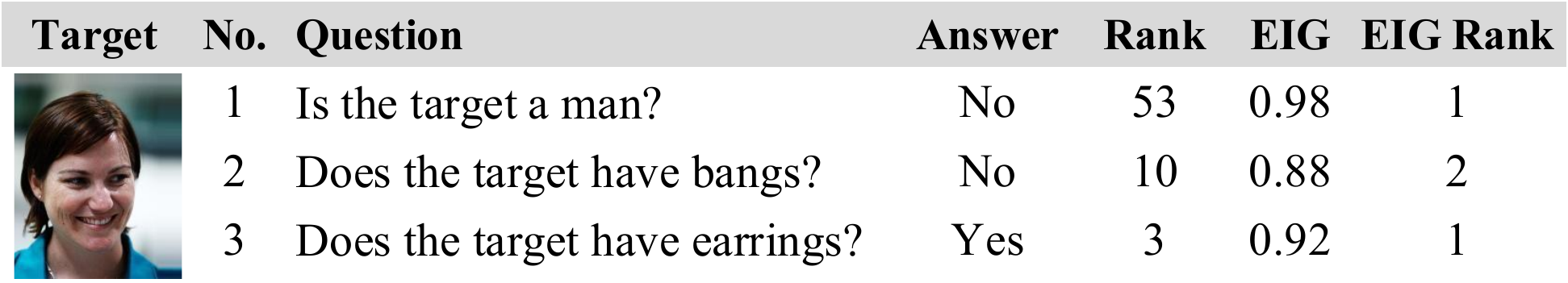}
        \caption{An example of progressive refinement in the CelebA image retrieval task, where each question improves the target's rank.}
        \label{fig:exp_female}
    \end{subfigure}
    \par\vspace{0.5em}
    \begin{subfigure}[b]{1\linewidth}
        \centering
        \includegraphics[page=4, width=1\linewidth]{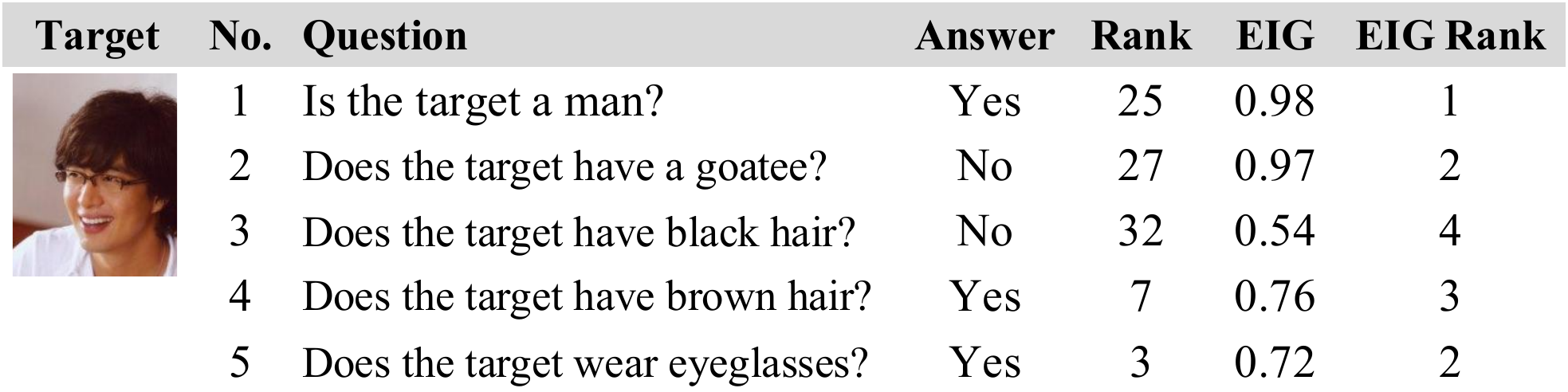}
        \caption{A more challenging retrieval task. The agent demonstrates resilience by recovering from an initial rank degradation caused by noisy similarity.}
        \label{fig:exp_male}
    \end{subfigure}
    \caption{\textbf{Examples from the image retrieval task.} For each turn, we show the target's rank, EIG of the chosen question, and its EIG Rank among all possible questions for the top-10 candidates.}
    \label{fig:combined_img_exp}
\end{figure}

\subsection{Results and Analysis}

\mypar{Tabular Retrieval Tasks.}
\Cref{tab:tabular_res} presents a performance comparison of SherlockLLM (Qwen2.5-7B + GRPO) against baseline models on tabular tasks. On Guess Number, the Oracle achieves a perfect SR in an optimal number of turns. SherlockLLM, while maintaining an almost perfect SR, yields an MT of 7.6. Compared to DeepSeek, SherlockLLM reduces MT by 5.0\%. On the Guess Who task, both SherlockLLM and DeepSeek attain a 100\% SR. Our policy requires an average of 6.25 turns per dialogue, a result that is on par with DeepSeek and one step more than the Oracle.
The Qwen2.5 models without fine-tuning underperform. The smallest variants (0.5B and 1.5B) fail to complete the tasks, indicating insufficient reasoning and instruction-following ability at this scale. Even the 32B variant stops at SR = 0.92/0.56 with MT = 9.39/11.44 on the two tasks. 
We compare the Qwen2.5-7B backbone against its SFT and GRPO counterparts. On Guess Number, SFT method increases SR by 29\% and reduces MT by 16.5\%. GRPO-tuned agent improves SR by 45.6\% and reduces MT by 38.7\%.  On Guess Who, SFT provides minimal improvement to the baseline model's SR and MT, showing improvements of 5.0\% and 12.1\%, respectively. For the GRPO method, improvements are even more striking: SR more than doubles from 0.46 to 1.0, with MT dropping by 52.9\%. 

\mypar{CelebA Image Retrieval.}
SherlockLLM achieves state-of-the-art performance on the CelebA image retrieval task, outperforming all baseline models (\cref{tab:img_res}). Compared to DeepSeek-V3.1, SherlockLLM shows a substantial advantage in dialogue efficiency, reducing MT by approximately 52\% and 38.8\% on the two test sets. In addition, its SR is significantly higher, showing improvements of 47.5\% and 91.7\% over DeepSeek-V3.1. 
Compared to the Qwen2.5-7B base model, the GRPO training yields remarkable gains: SR increases by 150\% and 200\% on the 100- and 500-image sets, respectively. The training also improves ranking performance, with MedR decreasing by 79.4\% and 92.3\%, and MR decreasing by 75.3\% and 60.9\%. Concurrently, dialogue efficiency is enhanced, as evidenced by a reduction in MT of 59.9\% and 43.3\% on the respective sets.

\mypar{Qualitative examples.}
In Guess Number and Guess Who (\cref{fig:combined_tabular_exp}), SherlockLLM learns near-optimal questioning strategies. For instance, in Guess Number, its approach is analogous to a binary search, consistently halving the candidate space to achieve a maximal EIG of 1.0. Similarly, in Guess Who, SherlockLLM learns to prioritize broad, discriminative questions about high-level attributes such as gender. These cases demonstrate that in well-defined problem spaces, our framework effectively guides the agent toward mathematically efficient policies. In the image retrieval, \Cref{fig:exp_female} shows a progressive refinement, where each subsequent question steadily improves the target’s rank. \cref{fig:exp_male} reveals SherlockLLM's ability to handle non-monotonic progression. In this challenging example, a negative answer temporarily worsens the target's rank from 25 to 32. This is a likely consequence of the complexities within the visual-semantic embedding space. However, SherlockLLM recovers from this setback, ultimately improving the target's rank to 3. This self-correction capability is a key advantage of our interactive approach over traditional one-shot query retrieval systems.

\begin{table}[t]
\centering
\caption{\textbf{Ablation study on different reward functions}.}
\label{tab:reward_func}
\begin{tabular}{lcc}
\toprule
\textbf{Reward Function} & \textbf{SR}$\uparrow$ & \textbf{MT}$\downarrow$ \\
\midrule
EIG Only & 0.99  & 7.55 \\
Step Penalty Only & 0.99  & \textbf{6.19} \\
EIG + Step Penalty & \textbf{1.0}  & 6.25 \\
\bottomrule
\end{tabular}
\end{table}

\mypar{Ablation Study on Reward Components.}
To ascertain the individual contribution of each component within our reward function (\cref{sec:reward}), we conduct an ablation study for the Guess Who task. We use three experimental conditions: 1) an agent rewarded solely based on EIG, 2) an agent guided by a terminal reward and a step penalty, and 3) an agent trained with the complete reward function. The success-contingent constant term is retained across all configurations.
The results are presented in \cref{tab:reward_func}. SherlockLLM trained with only the EIG reward learns to ask highly informative questions. However, this leads to a deliberate but slow strategy. SherlockLLM guided exclusively by a step penalty becomes highly efficient (6.19 MT), but optimized purely for speed, may not always be the most robust approach.
The combination of both components in the reward yields a more robust policy. This agent achieves a perfect success rate while maintaining a highly efficient average of 6.25 turns.
The per-step guidance from the EIG reward teaches the agent what constitutes an effective query, while the global step penalty incentivizes it to reach the goal expeditiously. This combination compels the agent to learn a strategy that is both effective and temporally efficient. 

\mypar{Impact of Retrieval Feedback.}
\Cref{tab:rag} isolates the effect of retrieval feedback. For Guess Who, injecting attribute-distribution feedback shortens the dialogue while maintaining ceiling success. This indicates that the agent already solves the task reliably, and feedback mainly helps it ask more informative follow-ups.
For the image retrieval, the choice of top-$K$ feedback is decisive. Moving from top-5 to top-10 retrieved images boosts SR by 36.4\%, and cuts MR by 62.0\%. Shrinking $K$ from 10 to 5 amplifies the noise and bias in the model's estimation of image feature distribution based on top-$K$ captions, leading to inaccurate expected information gain. Increasing $K$ can stabilize this distribution estimation, enabling the strategy to propose clearer and more valuable queries.

\begin{table}[t]
\centering
\caption{\textbf{Impact of retrieval feedback.} For the Guess Who, we compare a model with no feedback against one that receives the current attribute distribution. For the image retrieval, we compare the models using the top-5 or top-10 retrieved images.}
\label{tab:rag}
\setlength{\tabcolsep}{2pt}
\begin{tabular}{llcccc}
\toprule
\textbf{Task} & \textbf{Feedback} & \textbf{SR}$\uparrow$ & \textbf{MT}$\downarrow$ & \textbf{MedR}$\downarrow$ & \textbf{MR}$\downarrow$ \\
\midrule
\multirow{2}{*}{Guess Who}       & No Feedback & \textbf{1.0}  & 6.98 & - & - \\
                                 & Feedback    & \textbf{1.0}  & \textbf{6.25} & - & - \\
\hdashline
\multirow{2}{*}{Image Retrieval} & Top-5       & 0.66 & 9.92 & 4.0 & 16.7 \\
                                 & Top-10      & \textbf{0.90} & \textbf{6.15}  & \textbf{3.5}  & \textbf{6.35} \\
\bottomrule
\end{tabular}
\end{table}

\begin{table}[t]
\centering
\caption{\textbf{Comparison to different image retrieval backbones.}}
\label{tab:retrieval}
\setlength{\tabcolsep}{5pt}
\resizebox{\columnwidth}{!}{%
\begin{tabular}{lccccc}
\toprule
\textbf{Models} & \textbf{Data Size} & \textbf{SR}/\textbf{R@5}$\uparrow$ & \textbf{MT}$\downarrow$ & \textbf{MedR}$\downarrow$ & \textbf{MR}$\downarrow$ \\
\midrule
CLIP & 100 /  500 & 0.90 / \textbf{0.69} & 6.15  / \textbf{12.21} & 3.5 / \textbf{5.0} & 6.35 / \textbf{45.11} \\
BLIP & 100 / 500  & \textbf{1.0} / 0.27  & \textbf{5.22} / 20.44 & \textbf{2.0} / 64.0 & \textbf{2.36} / 112.11 \\
\bottomrule
\end{tabular}
}
\end{table}

\mypar{Impact of Retrieval Backbone.}
We investigate the impact of the retrieval backbone by comparing a training-free CLIP retriever and a fine-tuned BLIP \cite{li2022blip} counterpart. BLIP is first fine-tuned on a synthetic dialogue dataset (\cref{sec:blip}). The experiments are conducted on 100-image and 500-image set while holding the Questioner module and RL framework constant.
As shown in \cref{tab:retrieval}, the BLIP backbone achieves a higher ranking quality for the 100-image set.
However, this trend reversed in the 500-image setting. While CLIP's success rate remains stable at 0.69, BLIP's performance degrades to 0.27. A larger dataset demands longer dialogues and more compositional questions. The end-to-end BLIP does not generalize well, whereas the keyword-based CLIP demonstrates more robust scalability.

\section{Conclusions}
\label{sec:conclusions}

We propose a dialogue-driven information retrieval framework trained with reinforcement learning. Our method trains an agent to ask a sequence of optimally informative questions to efficiently narrow the search space. This approach aims to improve the retrieval accuracy and reduce the interaction turns. To validate our framework, we design three benchmark tasks across two data modalities: two structured data retrieval games and an image retrieval task. Experimental results demonstrate that our method significantly outperforms strong baselines across all tasks.

% \newpage 
% \input{sec/_1_format}
% \input{sec/_2_formatting}
% \input{sec/_3_finalcopy}
{
    \small
    \bibliographystyle{ieeenat_fullname}
    \bibliography{main}

\begin{thebibliography}{50}
\providecommand{\natexlab}[1]{#1}
\providecommand{\url}[1]{\texttt{#1}}
\expandafter\ifx\csname urlstyle\endcsname\relax
  \providecommand{\doi}[1]{doi: #1}\else
  \providecommand{\doi}{doi: \begingroup \urlstyle{rm}\Url}\fi

\bibitem[Ai et~al.(2023)Ai, Bai, Cao, Chang, Chen, Chen, Cheng, Dong, Dou, Feng, et~al.]{ai-InformationRetrievalMeets-2023}
Qingyao Ai, Ting Bai, Zhao Cao, Yi Chang, Jiawei Chen, Zhumin Chen, Zhiyong Cheng, Shoubin Dong, Zhicheng Dou, Fuli Feng, et~al.
\newblock Information retrieval meets large language models: a strategic report from chinese ir community.
\newblock \emph{AI open}, 2023.

\bibitem[Aliannejadi et~al.(2019)Aliannejadi, Zamani, Crestani, and Croft]{aliannejadi2019asking}
Mohammad Aliannejadi, Hamed Zamani, Fabio Crestani, and W~Bruce Croft.
\newblock Asking clarifying questions in open-domain information-seeking conversations.
\newblock In \emph{SIGIR}, 2019.

\bibitem[Borgeaud et~al.(2022)Borgeaud, Mensch, Hoffmann, Cai, Rutherford, Millican, Van Den~Driessche, Lespiau, Damoc, Clark, et~al.]{borgeaud2022improving}
Sebastian Borgeaud, Arthur Mensch, Jordan Hoffmann, Trevor Cai, Eliza Rutherford, Katie Millican, George~Bm Van Den~Driessche, Jean-Baptiste Lespiau, Bogdan Damoc, Aidan Clark, et~al.
\newblock Improving language models by retrieving from trillions of tokens.
\newblock In \emph{International conference on machine learning}, pages 2206--2240. PMLR, 2022.

\bibitem[Cao et~al.(2025)Cao, Zhao, Cheng, Shu, Chen, Liu, Liang, Zhao, Yan, and Li]{10766898}
Yuji Cao, Huan Zhao, Yuheng Cheng, Ting Shu, Yue Chen, Guolong Liu, Gaoqi Liang, Junhua Zhao, Jinyue Yan, and Yun Li.
\newblock Survey on large language model-enhanced reinforcement learning: Concept, taxonomy, and methods.
\newblock In \emph{IEEE Transactions on Neural Networks and Learning Systems}, 2025.

\bibitem[Chen et~al.(2024)Chen, Zhang, Dong, Zhou, and Wang]{chen2024enhancing}
Peiyuan Chen, Zecheng Zhang, Yiping Dong, Li Zhou, and Han Wang.
\newblock Enhancing visual question answering through ranking-based hybrid training and multimodal fusion.
\newblock \emph{arXiv preprint arXiv:2408.07303}, 2024.

\bibitem[Cormack et~al.(2009)Cormack, Clarke, and Buettcher]{cormack2009reciprocal}
Gordon~V Cormack, Charles~LA Clarke, and Stefan Buettcher.
\newblock Reciprocal rank fusion outperforms condorcet and individual rank learning methods.
\newblock In \emph{SIGIR}, 2009.

\bibitem[Dai et~al.(2022)Dai, Chaganty, Zhao, Amini, Rashid, Green, and Guu]{dai2022dialog}
Zhuyun Dai, Arun~Tejasvi Chaganty, Vincent~Y Zhao, Aida Amini, Qazi~Mamunur Rashid, Mike Green, and Kelvin Guu.
\newblock Dialog inpainting: Turning documents into dialogs.
\newblock In \emph{International conference on machine learning}, pages 4558--4586. PMLR, 2022.

\bibitem[Datta et~al.(2008)Datta, Joshi, Li, and Wang]{datta-ImageRetrievalIdeas-2008}
Ritendra Datta, Dhiraj Joshi, Jia Li, and James~Z. Wang.
\newblock Image retrieval: Ideas, influences, and trends of the new age.
\newblock In \emph{ACM Comput. Surv.}, 2008.

\bibitem[Guo et~al.(2025)Guo, Yang, Zhang, Song, Zhang, Xu, Zhu, Ma, Wang, Bi, et~al.]{guo2025deepseek}
Daya Guo, Dejian Yang, Haowei Zhang, Junxiao Song, Ruoyu Zhang, Runxin Xu, Qihao Zhu, Shirong Ma, Peiyi Wang, Xiao Bi, et~al.
\newblock Deepseek-r1: Incentivizing reasoning capability in llms via reinforcement learning.
\newblock \emph{arXiv}, 2025.

\bibitem[Guu et~al.(2020)Guu, Lee, Tung, Pasupat, and Chang]{guu2020retrieval}
Kelvin Guu, Kenton Lee, Zora Tung, Panupong Pasupat, and Mingwei Chang.
\newblock Retrieval augmented language model pre-training.
\newblock In \emph{International conference on machine learning}, pages 3929--3938. PMLR, 2020.

\bibitem[Hu et~al.(2022)Hu, Shen, Wallis, Allen-Zhu, Li, Wang, Wang, Chen, et~al.]{hu2022lora}
Edward~J Hu, Yelong Shen, Phillip Wallis, Zeyuan Allen-Zhu, Yuanzhi Li, Shean Wang, Lu Wang, Weizhu Chen, et~al.
\newblock Lora: Low-rank adaptation of large language models.
\newblock In \emph{ICLR}, 2022.

\bibitem[Hu et~al.(2018)Hu, Wu, Luo, Tao, Xu, Wu, and Chen]{hu2018playing}
Huang Hu, Xianchao Wu, Bingfeng Luo, Chongyang Tao, Can Xu, Wei Wu, and Zhan Chen.
\newblock Playing 20 question game with policy-based reinforcement learning.
\newblock In \emph{EMNLP}, 2018.

\bibitem[Hu et~al.(2024)Hu, Liu, Feng, Zhao, Ng, Luu, He, Koh, and Hooi]{hu_uncertainty_2024}
Zhiyuan Hu, Chumin Liu, Xidong Feng, Yilun Zhao, See-Kiong Ng, Anh~Tuan Luu, Junxian He, Pang~Wei Koh, and Bryan Hooi.
\newblock Uncertainty of {Thoughts}: {Uncertainty}-{Aware} {Planning} {Enhances} {Information} {Seeking} in {LLMs}.
\newblock In \emph{NeurIPS}, 2024.

\bibitem[Hurst et~al.(2024)Hurst, Lerer, Goucher, Perelman, Ramesh, Clark, Ostrow, Welihinda, Hayes, Radford, et~al.]{hurst2024gpt}
Aaron Hurst, Adam Lerer, Adam~P Goucher, Adam Perelman, Aditya Ramesh, Aidan Clark, AJ Ostrow, Akila Welihinda, Alan Hayes, Alec Radford, et~al.
\newblock Gpt-4o system card.
\newblock In \emph{arXiv preprint arXiv:2410.21276}, 2024.

\bibitem[Izacard et~al.(2023)Izacard, Lewis, Lomeli, Hosseini, Petroni, Schick, Dwivedi-Yu, Joulin, Riedel, and Grave]{izacard2023atlas}
Gautier Izacard, Patrick Lewis, Maria Lomeli, Lucas Hosseini, Fabio Petroni, Timo Schick, Jane Dwivedi-Yu, Armand Joulin, Sebastian Riedel, and Edouard Grave.
\newblock Atlas: Few-shot learning with retrieval augmented language models.
\newblock \emph{Journal of Machine Learning Research}, 24\penalty0 (251):\penalty0 1--43, 2023.

\bibitem[Jagerman et~al.(2023)Jagerman, Zhuang, Qin, Wang, and Bendersky]{jagerman2023query}
Rolf Jagerman, Honglei Zhuang, Zhen Qin, Xuanhui Wang, and Michael Bendersky.
\newblock Query expansion by prompting large language models.
\newblock \emph{arXiv preprint arXiv:2305.03653}, 2023.

\bibitem[Jiang et~al.(2025)Jiang, Lin, Cao, Tian, Kang, Wang, Sun, and Han]{jiang2025deepretrieval}
Pengcheng Jiang, Jiacheng Lin, Lang Cao, Runchu Tian, SeongKu Kang, Zifeng Wang, Jimeng Sun, and Jiawei Han.
\newblock Deepretrieval: Hacking real search engines and retrievers with large language models via reinforcement learning.
\newblock \emph{CoRR}, 2025.

\bibitem[Karpukhin et~al.(2020)Karpukhin, Oguz, Min, Wu, Edunov, Chen, and Yih]{karpukhin2020dense}
Vladimir Karpukhin, Barlas Oguz, Sewon Min, Ledell Wu, Sergey Edunov, Danqi Chen, and Wen-tau Yih.
\newblock Dense passage retrieval for open-domain question answering.
\newblock In \emph{EMNLP}, 2020.

\bibitem[Keh et~al.(2023)Keh, Chiu, and Fried]{keh_asking_2024}
Sedrick Keh, Justin~T Chiu, and Daniel Fried.
\newblock Asking more informative questions for grounded retrieval.
\newblock In \emph{Findings of the Association for Computational Linguistics: NAACL 2024}, 2023.

\bibitem[Lee et~al.(2024)Lee, Yu, Park, Yi, and Yoon]{lee-etal-2024-interactive}
Saehyung Lee, Sangwon Yu, Junsung Park, Jihun Yi, and Sungroh Yoon.
\newblock Interactive text-to-image retrieval with large language models: A plug-and-play approach.
\newblock In \emph{62nd Annual Meeting of the Association for Computational Linguistics}, 2024.

\bibitem[Levy et~al.(2023)Levy, Ben-Ari, Darshan, and Lischinski]{levy2023chatting}
Matan Levy, Rami Ben-Ari, Nir Darshan, and Dani Lischinski.
\newblock Chatting makes perfect: Chat-based image retrieval.
\newblock In \emph{NeurIPS}, 2023.

\bibitem[Li et~al.(2022)Li, Li, Xiong, and Hoi]{li2022blip}
Junnan Li, Dongxu Li, Caiming Xiong, and Steven Hoi.
\newblock Blip: Bootstrapping language-image pre-training for unified vision-language understanding and generation.
\newblock In \emph{PMLR}, 2022.

\bibitem[Li et~al.(2024)Li, Balachandran, Feng, Ilgen, Pierson, Koh, and Tsvetkov]{li2024mediq}
Stella Li, Vidhisha Balachandran, Shangbin Feng, Jonathan Ilgen, Emma Pierson, Pang Wei~W Koh, and Yulia Tsvetkov.
\newblock Mediq: Question-asking llms and a benchmark for reliable interactive clinical reasoning.
\newblock In \emph{NeurIPS}, 2024.

\bibitem[Liu et~al.(2024)Liu, Feng, Xue, Wang, Wu, Lu, Zhao, Deng, Zhang, Ruan, et~al.]{liu2024deepseek}
Aixin Liu, Bei Feng, Bing Xue, Bingxuan Wang, Bochao Wu, Chengda Lu, Chenggang Zhao, Chengqi Deng, Chenyu Zhang, Chong Ruan, et~al.
\newblock Deepseek-v3 technical report.
\newblock \emph{arXiv preprint arXiv:2412.19437}, 2024.

\bibitem[Liu et~al.(2015)Liu, Luo, Wang, and Tang]{liu2015faceattributes}
Ziwei Liu, Ping Luo, Xiaogang Wang, and Xiaoou Tang.
\newblock Deep learning face attributes in the wild.
\newblock In \emph{ICCV}, 2015.

\bibitem[Loshchilov and Hutter(2017)]{loshchilov2017decoupled}
Ilya Loshchilov and Frank Hutter.
\newblock Decoupled weight decay regularization.
\newblock In \emph{arXiv preprint arXiv:1711.05101}, 2017.

\bibitem[Ma et~al.(2023)Ma, Gong, He, Zhao, and Duan]{ma2023query}
Xinbei Ma, Yeyun Gong, Pengcheng He, Hai Zhao, and Nan Duan.
\newblock Query rewriting in retrieval-augmented large language models.
\newblock In \emph{EMNLP}, 2023.

\bibitem[Mao et~al.(2023)Mao, Dou, Mo, Hou, Chen, and Qian]{mao-etal-2023-large}
Kelong Mao, Zhicheng Dou, Fengran Mo, Jiewen Hou, Haonan Chen, and Hongjin Qian.
\newblock Large language models know your contextual search intent: A prompting framework for conversational search.
\newblock In \emph{EMNLP}, 2023.

\bibitem[Mao et~al.(2024)Mao, Jiang, Chen, Li, Wang, Wang, Xie, Huang, Chen, and Zhang]{mao2024rafe}
Shengyu Mao, Yong Jiang, Boli Chen, Xiao Li, Peng Wang, Xinyu Wang, Pengjun Xie, Fei Huang, Huajun Chen, and Ningyu Zhang.
\newblock Rafe: Ranking feedback improves query rewriting for rag.
\newblock In \emph{EMNLP}, 2024.

\bibitem[Mazzaccara et~al.(2024)Mazzaccara, Testoni, and Bernardi]{mazzaccara-etal-2024-learning}
Davide Mazzaccara, Alberto Testoni, and Raffaella Bernardi.
\newblock Learning to ask informative questions: Enhancing {LLM}s with preference optimization and expected information gain.
\newblock In \emph{Findings of the Association for Computational Linguistics: EMNLP 2024}, 2024.

\bibitem[Peng et~al.(2024)Peng, Li, Jiang, Wang, Ou, Zeng, Xu, Xu, and Chen]{peng2024large}
Wenjun Peng, Guiyang Li, Yue Jiang, Zilong Wang, Dan Ou, Xiaoyi Zeng, Derong Xu, Tong Xu, and Enhong Chen.
\newblock Large language model based long-tail query rewriting in taobao search.
\newblock In \emph{ACM}, 2024.

\bibitem[Rackauckas(2024)]{rackauckas2024rag}
Zackary Rackauckas.
\newblock Rag-fusion: a new take on retrieval-augmented generation.
\newblock \emph{arXiv preprint arXiv:2402.03367}, 2024.

\bibitem[Radford et~al.(2021)Radford, Kim, Hallacy, Ramesh, Goh, Agarwal, Sastry, Askell, Mishkin, Clark, et~al.]{radford2021learning}
Alec Radford, Jong~Wook Kim, Chris Hallacy, Aditya Ramesh, Gabriel Goh, Sandhini Agarwal, Girish Sastry, Amanda Askell, Pamela Mishkin, Jack Clark, et~al.
\newblock Learning transferable visual models from natural language supervision.
\newblock In \emph{ICML}, 2021.

\bibitem[Rafailov et~al.(2023)Rafailov, Sharma, Mitchell, Manning, Ermon, and Finn]{rafailov2023direct}
Rafael Rafailov, Archit Sharma, Eric Mitchell, Christopher~D Manning, Stefano Ermon, and Chelsea Finn.
\newblock Direct preference optimization: Your language model is secretly a reward model.
\newblock In \emph{NeurIPS}, 2023.

\bibitem[Robertson et~al.(2009)Robertson, Zaragoza, et~al.]{robertson2009probabilistic}
Stephen Robertson, Hugo Zaragoza, et~al.
\newblock The probabilistic relevance framework: Bm25 and beyond.
\newblock \emph{Foundations and Trends in Information Retrieval}, 2009.

\bibitem[Salton and McGill(1986)]{salton-IntroductionModernInformation-1986}
Gerard Salton and Michael~J. McGill.
\newblock \emph{Introduction to Modern Information Retrieval}.
\newblock McGraw-Hill, Inc., USA, 1986.

\bibitem[Salton et~al.(1975)Salton, Wong, and Yang]{salton1975vector}
Gerard Salton, Anita Wong, and Chung-Shu Yang.
\newblock A vector space model for automatic indexing.
\newblock \emph{Communications of the ACM}, 1975.

\bibitem[Shannon(1948)]{shannon1948mathematical}
Claude~E Shannon.
\newblock A mathematical theory of communication.
\newblock In \emph{The Bell system technical journal}, 1948.

\bibitem[Shao et~al.(2024)Shao, Wang, Zhu, Xu, Song, Bi, Zhang, Zhang, Li, Wu, and Guo]{shao2024deepseekmath}
Zhihong Shao, Peiyi Wang, Qihao Zhu, Runxin Xu, Junxiao Song, Xiao Bi, Haowei Zhang, Mingchuan Zhang, Y.~K. Li, Y. Wu, and Daya Guo.
\newblock Deepseekmath: Pushing the limits of mathematical reasoning in open language models.
\newblock In \emph{arXiv preprint arXiv:2402.03300}, 2024.

\bibitem[Siro et~al.(2024)Siro, Yuan, Aliannejadi, and de~Rijke]{siro2024agent}
Clemencia Siro, Yifei Yuan, Mohammad Aliannejadi, and Maarten de Rijke.
\newblock Agent-cq: Automatic generation and evaluation of clarifying questions for conversational search with llms.
\newblock In \emph{arXiv preprint arXiv:2410.19692}, 2024.

\bibitem[Srinivasan et~al.(2022)Srinivasan, Raman, Samanta, Liao, Bertelli, and Bendersky]{srinivasan2022quill}
Krishna Srinivasan, Karthik Raman, Anupam Samanta, Lingrui Liao, Luca Bertelli, and Michael Bendersky.
\newblock Quill: Query intent with large language models using retrieval augmentation and multi-stage distillation.
\newblock In \emph{Proceedings of the 2022 Conference on Empirical Methods in Natural Language Processing: Industry Track}, 2022.

\bibitem[Team(2024)]{qwen2.5}
Qwen Team.
\newblock Qwen2.5 technical report.
\newblock In \emph{arXiv preprint arXiv:2412.15115}, 2024.

\bibitem[Wang et~al.(2023)Wang, Yang, and Wei]{wang2023query2doc}
Liang Wang, Nan Yang, and Furu Wei.
\newblock Query2doc: Query expansion with large language models.
\newblock In \emph{EMNLP}, 2023.

\bibitem[White et~al.(2021)White, Poesia, Hawkins, Sadigh, and Goodman]{white-etal-2021-open}
Julia White, Gabriel Poesia, Robert Hawkins, Dorsa Sadigh, and Noah Goodman.
\newblock Open-domain clarification question generation without question examples.
\newblock In \emph{EMNLP}, 2021.

\bibitem[Wilson et~al.(2025)Wilson, Carter, and Graham]{wilson-ContextualizingSearchQueries-2025}
Raymond Wilson, Chase Carter, and Cole Graham.
\newblock Contextualizing search queries {{In-context}} learning for conversational rewriting with {{LLMs}}.
\newblock In \emph{arXiv preprint arXiv:2502.15009}, 2025.

\bibitem[Young et~al.(2014)Young, Lai, Hodosh, and Hockenmaier]{young-etal-2014-image}
Peter Young, Alice Lai, Micah Hodosh, and Julia Hockenmaier.
\newblock From image descriptions to visual denotations: New similarity metrics for semantic inference over event descriptions.
\newblock In \emph{Transactions of the Association for Computational Linguistics}, 2014.

\bibitem[Yu et~al.(2020)Yu, Chen, Wang, Lei, and Artzi]{yu-etal-2020-interactive}
Lili Yu, Howard Chen, Sida~I. Wang, Tao Lei, and Yoav Artzi.
\newblock Interactive classification by asking informative questions.
\newblock In \emph{58th Annual Meeting of the Association for Computational Linguistics}, 2020.

\bibitem[Yuan et~al.(2025)Yuan, Abbasiantaeb, Aliannejadi, and Deng]{yuan2025query}
Yifei Yuan, Zahra Abbasiantaeb, Mohammad Aliannejadi, and Yang Deng.
\newblock Query understanding in llm-based conversational information seeking.
\newblock In \emph{Proceedings of the 48th International ACM SIGIR Conference on Research and Development in Information Retrieval}, pages 4098--4101, 2025.

\bibitem[Zamani et~al.(2020)Zamani, Mitra, Chen, Lueck, Diaz, Bennett, Craswell, and Dumais]{zamani2020analyzing}
Hamed Zamani, Bhaskar Mitra, Everest Chen, Gord Lueck, Fernando Diaz, Paul~N Bennett, Nick Craswell, and Susan~T Dumais.
\newblock Analyzing and learning from user interactions for search clarification.
\newblock In \emph{SIGIR}, 2020.

\bibitem[Zhu et~al.(2025)Zhu, Yuan, Wang, Liu, Liu, Deng, Chen, Liu, Dou, and Wen]{zhu-LargeLanguageModels-2025}
Yutao Zhu, Huaying Yuan, Shuting Wang, Jiongnan Liu, Wenhan Liu, Chenlong Deng, Haonan Chen, Zheng Liu, Zhicheng Dou, and Ji-Rong Wen.
\newblock Large language models for information retrieval: A survey.
\newblock \emph{ACM Trans. Inf. Syst.}, 2025.

\end{thebibliography}
}
\appendix
\newpage
\clearpage
\setcounter{page}{1}
\maketitlesupplementary

\section{Analysis of Alternative Methods}

This section justifies the core design choices of the SherlockLLM for image retrieval. First, we justify our use of a rank-based reward function, which we compare to an alternative similarity-based formulation. Second, we evaluate our ranking algorithm. This evaluation involves a comparison with other common fusion methods.

\subsection{Image Retrieval Reward Signal.}
To optimize the training of our agent for the interactive image retrieval task, we investigated the efficacy of two different reward signal formulations. Intuitively, if the similarity score between the target and the query is higher, then it will rank higher. The objective of this experiment was to determine whether a direct optimization of the target's rank is more effective than using an implicit signal based on text-image similarity. All comparative experiments were trained on the 100-image dataset. Except for the definition of the reward function, all other conditions and training procedures remained consistent.

As \Cref{tab:ranking_signal} shows, both reward signals achieve an identical SR of 0.90. This indicates that both formulations are feasible. However, we observe clear differences in efficiency and ranking quality. The agent trained with the Rank-based reward demonstrates superior efficiency. It requires an MT of 6.15, a 6.1\% reduction in conversational length compared to the 6.55 turns required by the agent trained on the Similarity score. Furthermore, the Rank-based reward achieves a MedR of 3.5, representing a 12.5\% improvement for the Similarity-based reward.

Interestingly, the Similarity-based reward achieves a slightly better MR of 5.9. The Rank-based reward's MR is approximately 7.6\% worse by comparison. We hypothesize that while the similarity-based agent is less efficient on average, it generates more exceptionally descriptive questions that result in a perfect R@1 retrieval. These highly successful instances would significantly pull down the mean rank. Conversely, the rank-based agent is more consistent, reliably placing the target in the top few positions as suggested by its superior median rank, but with fewer perfect R@1 retrievals. This is substantiated by an analysis of the R@1 metric. The agent trained on Similarity achieves an R@1 of 0.25, whereas the agent trained on Rank only reaches 0.1.

Despite the superior mean rank achieved with the similarity signal, we select the Rank-based reward as the main method in this work. The significant gains in dialogue efficiency and more reliable typical performance (12.5\% better median rank) are of paramount importance for an interactive retrieval system. 

\begin{table}[t]
\centering
\caption{Comparison of retrieval performance on CelebA image retrieval task using Rank vs. Similarity as the reward signal.}
\label{tab:ranking_signal}
\resizebox{\columnwidth}{!}{%
\begin{tabular}{lccccc}
\toprule
\textbf{Reward Signal} & \textbf{SR/R@5}$\uparrow$  &  \textbf{MT}$\downarrow$  &  \textbf{MedR}$\downarrow$ & \textbf{MR}$\downarrow$ & \textbf{R@1}$\uparrow$ \\
\midrule
Rank        & \textbf{0.90} & \textbf{6.15} & \textbf{3.5} & 6.35         & 0.1  \\
Similarity  & \textbf{0.90} & 6.55          & 4.0          & \textbf{5.9} & \textbf{0.25} \\
\bottomrule
\end{tabular}
}
\end{table}

\subsection{Ranking Algorithm Comparison}
The ranking algorithm is a critical component of our agent for the image retrieval task. It translates the dialogue history into a ranked list of candidate images at each turn. We justify our choice of algorithm by comparing our proposed method against two plausible and computationally efficient alternatives.

Our primary ranking algorithm, detailed in \Cref{alg:clip_retriever}, operates on a cumulative basis. This method is designed for high precision because it leverages all information gathered up to the current turn. The main drawback of this approach is its computational cost. Recalculating similarities for the entire keyword history at every turn can introduce significant latency. We therefore design two alternative algorithms that improve efficiency by avoiding these historical recalculations.

\paragraph{Reciprocal Rank Fusion (RRF).} RRF \cite{cormack2009reciprocal} is an algorithm that combines multiple ranked lists into a single, more optimal result. This method has seen wide application in modern retrieval systems \cite{rackauckas2024rag, chen2024enhancing}. The algorithm is entirely training-free. It operates only on the rank positions of items and does not require their underlying relevance scores. The RRF score for a candidate is the sum of the reciprocal of its rank across all lists. A small constant $k = 60$ is typically added to the denominator. The formula is given by the following equation.

\begin{equation}
    \label{eq:rrf}
    \operatorname{Score}_{RRF}(c)=\sum_{i=1}^{N} \frac{1}{k+\operatorname{rank}_{i}(c)}
\end{equation}

\noindent where $\operatorname{rank}_{i}(c)$ is the rank of candidate $c$ in the $i$-th list from a total of $N$ lists.

In our implementation of this alternative, we use RRF to fuse information sequentially across dialogue turns. At each turn $t$, we apply \Cref{alg:clip_retriever} using only the most recent keywords. This process generates a turn-specific ranked list, denoted as $\operatorname{Rank}_{t} = \{\operatorname{rank}_{1},\operatorname{rank}_{2},\dots,\operatorname{rank}_{t}\}$. The consolidated ranking for the current turn is then produced by applying \Cref{eq:rrf} to the complete set of lists generated up to that point. This approach allows for turn-by-turn updates without the computational overhead of re-evaluating the entire dialogue history.

\paragraph{Sequential Score Multiplication (SSM).} SSM also operates sequentially but fuses information at the score level. This method functions as a soft logical AND gate. A candidate image must satisfy the constraints from every turn to remain a viable option.

At each turn $t$, a turn-specific score is calculated for each candidate image based on its similarity to the current turn's positive keywords ($Sim_{positive}$) and negative keywords ($Sim_{negative}$), with both values scaled to a $[0, 1]$ range. The turn score is then defined by:

\begin{equation}
    \label{eq:ssm}
    Score_t = Sim_{positive} \times (1 - Sim_{negative})
\end{equation}

The final relevance score for an image at turn t is the cumulative product of all turn scores up to that point, where $S_{final, t} = S_{final, t-1} \times Score_t$. The initial score is set to $S_{final, 0} = 1$.  This cumulative multiplication ensures that only images that consistently receive high scores across all turns will maintain a high final ranking.

\paragraph{Results.} We conduct the experiment on the 100-image dataset. The dialogues are generated for each target image using the methodology described in \Cref{sec:blip}.

As shown in \Cref{tab:ranking_algori}, the results demonstrate the clear superiority of our proposed algorithm across all key metrics. Our SherlockLLM algorithm achieves an R@5 of 0.42, a R@10 of 0.57, and a MedR of 7.5. In contrast, the performance of the two methods is significantly worse. The RRF algorithm's R@5 is only 0.06, while the SSM algorithm fails to rank any target image in the top position. Their Median and Mean Ranks both exceed 50. This result indicates the target image did not even enter a meaningful ranking range.

The failure of the RRF method stems from its incorrect assumption that information from each turn constitutes parallel and independent evidence. In our interactive scenario, the dialogue imposes sequential, conditional constraints. For example, a query for ``red hair" after a positive response to ``is it a man?" implies a search for a ``man with red hair." The RRF fusion mechanism cannot model this implicit AND condition. Instead, it simply combines the ranked list for ``man" with the ranked list for ``red hair," failing to properly constrain the search space as the dialogue progresses.

The SSM method fails because it operates on the flawed assumption that similarity scores from each turn are equally important and directly comparable. However, query terms vary in their semantics and difficulty, which leads to vastly different score distributions from the retrieval model. Multiplying these scores, which are derived from different scales, distorts the true relevance of candidates and produces unreliable rankings.

\begin{table}[t]
\centering
\caption{Comparison to different image retrieval ranking algorithms.}
\label{tab:ranking_algori}
\resizebox{\columnwidth}{!}{%
\begin{tabular}{lccccc}
\toprule
\textbf{Algorithm} & \textbf{R@1}$\uparrow$& \textbf{R@5}$\uparrow$& \textbf{R@10}$\uparrow$ & \textbf{MedR}$\downarrow$ & \textbf{MR}$\downarrow$ \\
\midrule
SherlockLLM     & \textbf{0.15} & \textbf{0.42} & \textbf{0.57} & \textbf{7.5} & \textbf{15.33} \\
RRF             & 0.02 & 0.06 & 0.1 & 50.5 & 50.41 \\
SSM  & 0.0 & 0.01 & 0.05 & 53.5 & 53.73 \\
\bottomrule
\end{tabular}
}
\end{table}

\section{Additional Case Studies}

\subsection{Guess Number}
\begin{figure*}[h!]
  \centering
   \includegraphics[page=1, width=0.9\linewidth]{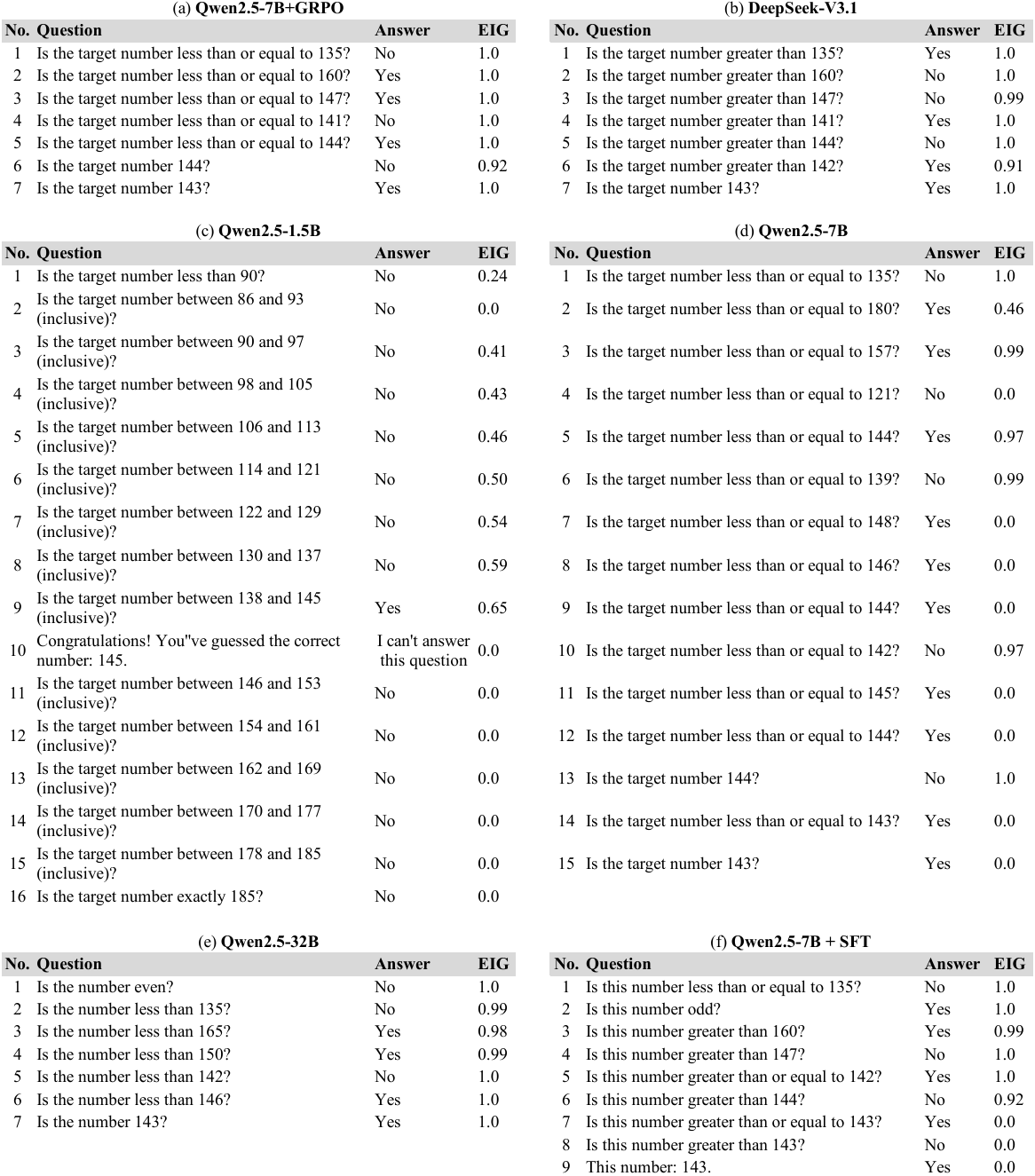}
   \caption{Examples of the Guess Number performed by baseline models and SherlockLLM.}
   \label{fig:guessnum_exp}
\end{figure*}

\Cref{fig:guessnum_exp} shows examples of three models on the Guess Number task. The goal is to find the target number 143 within an integer set $\{86,\; \dots,\; 185\}$. The DeepSeek and Qwen2.5-32B execute a near-optimal binary search. In contrast, Qwen2.5-1.5B performs an inefficient linear scan by sequentially checking fixed-size intervals. While it correctly identifies the target's range in turn 9, the model fails to narrow the search further and exceeds the maximum turn limit. 

The Qwen2.5-7B attempts a binary search but demonstrates a flawed strategy. The model commits severe logical errors and repetitive questions. This indicates that 7B can imitate the form of an optimal policy, but lacks the robust reasoning for the task. 

SFT agent learns to mimic actions from the training data, but it fails to optimize for turn efficiency. In the initial phase (turns 1-6), the agent's performance is optimal. It consistently poses a series of questions with high or perfect EIG scores. However, the agent's strategy falters in the endgame phase. At turn 7, instead of making a decisive guess, the agent asks other questions.

\subsection{Guess Who}

\begin{figure*}[h!]
  \centering
   \includegraphics[page=1, width=0.9\linewidth]{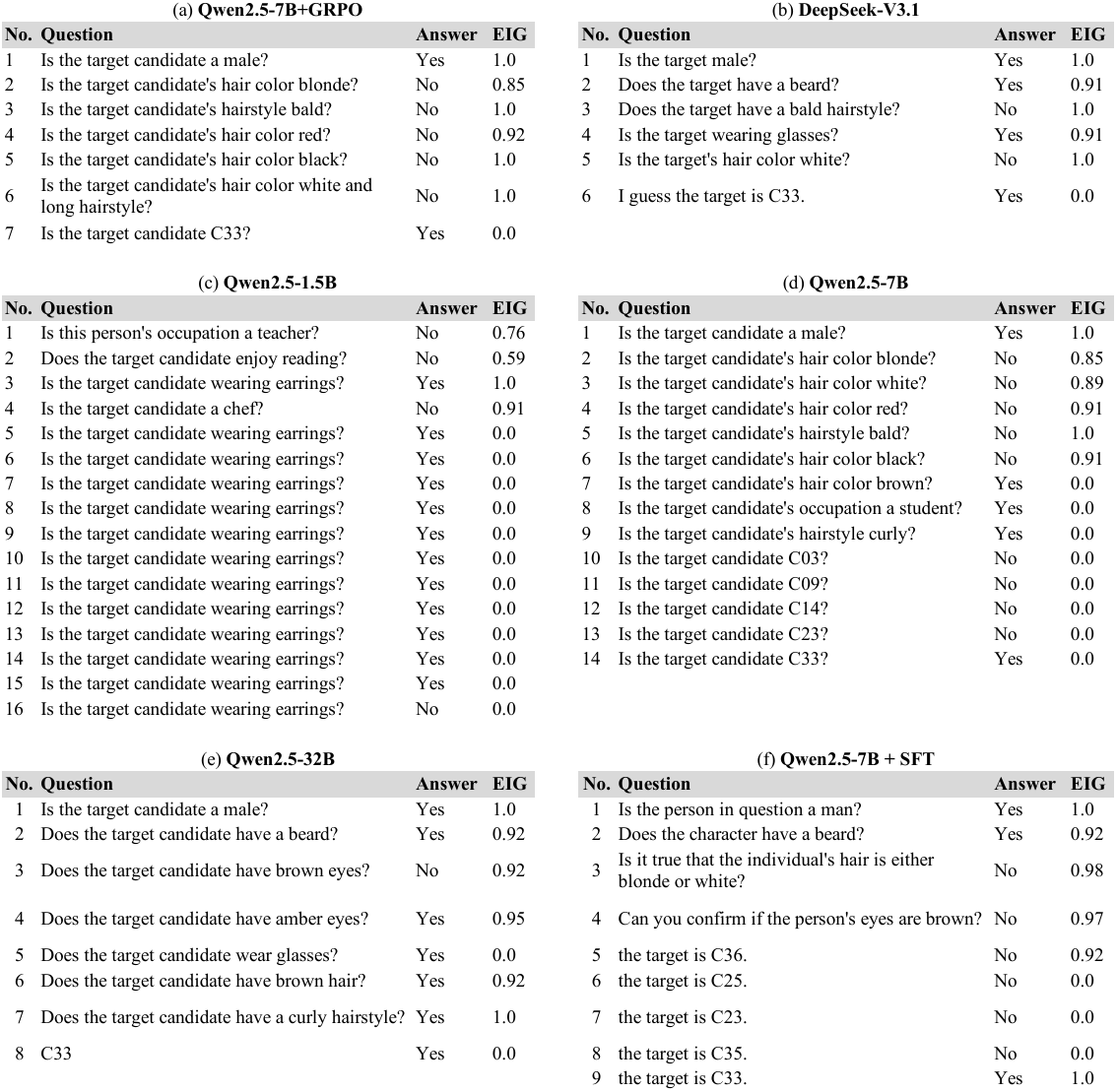}
   \caption{Examples of the Guess Who task performed by baseline models and SherlockLLM. Each example presents the full dialogue history for identifying the target C33.}
   \label{fig:guesswho_exp}
\end{figure*}

\Cref{fig:guesswho_exp} presents a comparison of the dialogue strategies that the three models employed on the Guess Who task. The DeepSeek demonstrates a logical and efficient strategy. It queries attributes step by step and consistently selects questions with a high EIG. In contrast, the Qwen2.5-1.5B exhibits a flawed policy. It enters a repetitive loop, repeatedly asking a question with zero EIG. This behavior indicates a deficiency in multi-turn conversation capabilities and ultimately results in task failure. 

The Qwen2.5-7B model shows an improvement over its 1.5B counterpart. It initially asks general attributes questions. However, while target C33 was the only remaining candidate after turn 6, the model proceeded with several rounds of unnecessary guessing before making the correct identification. 

Qwen2.5-32B is better than Qwen2.5-7B. It asks high-EIG questions about character attributes. However, its strategy is demonstrably suboptimal and contains clear logical inconsistencies. It fails to perform the necessary logical deduction on the constrained candidate pool.

Similar to the behavior of SFT in Guess Number task, SFT agent can mimic questioning, but its policy is weak when the optimal action shifts from information gathering to logical inference.

\subsection{CelebA Image Retrieval}

\begin{figure}
    \centering
   \begin{subfigure}[b]{1\linewidth}
        \centering
        \includegraphics[page=1, width=0.9\linewidth]{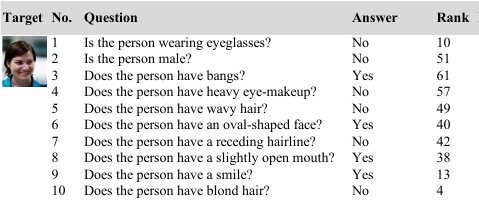}
        \caption{DeepSeek model.}
        \label{fig:image_exp_deepseek_sub}     
    \end{subfigure}
    \par\vspace{0.5em} 
    \begin{subfigure}[b]{1\linewidth}
        \centering
        \includegraphics[page=2, width=0.9\linewidth]{figs/ex_image.pdf}
        \caption{Qwen2.5-1.5B model.}
        \label{fig:image_exp_qwen1.5b_sub}
    \end{subfigure}
    \par\vspace{0.5em} 
    \begin{subfigure}[b]{1\linewidth}
        \centering
        \includegraphics[page=3, width=0.9\linewidth]{figs/ex_image.pdf}
        \caption{Qwen2.5-7B model.}
        \label{fig:image_exp_qwen7b_sub} 
    \end{subfigure}
    \caption{Examples of three models on the CelebA image retrieval task.}
    \label{fig:img_exp_comparison}
\end{figure}

Figure \cref{fig:img_exp_comparison} illustrates the model behaviors on the CelebA image retrieval task. DeepSeek uses a diverse and exploratory strategy, similar to its performance on the Guess Who task. It successfully completes the task by querying a wide range of attributes. In contrast, Qwen2.5-1.5B exhibits a failure by generating severe hallucinations after several initial turns, repeatedly asking for attributes that do not exist in the dataset, such as ``tattoos" and ``scars". Qwen2.5-7B demonstrates a distinct failure mode: it continuously asks the same questions. This repetitive behavior halts the retrieval process, ultimately leading to task failure.

\section{Guess Who Dataset}
\label{sec:guesswho_dataset}

\begin{figure}[t]
  \centering
   \includegraphics[page=1, width=1\linewidth]{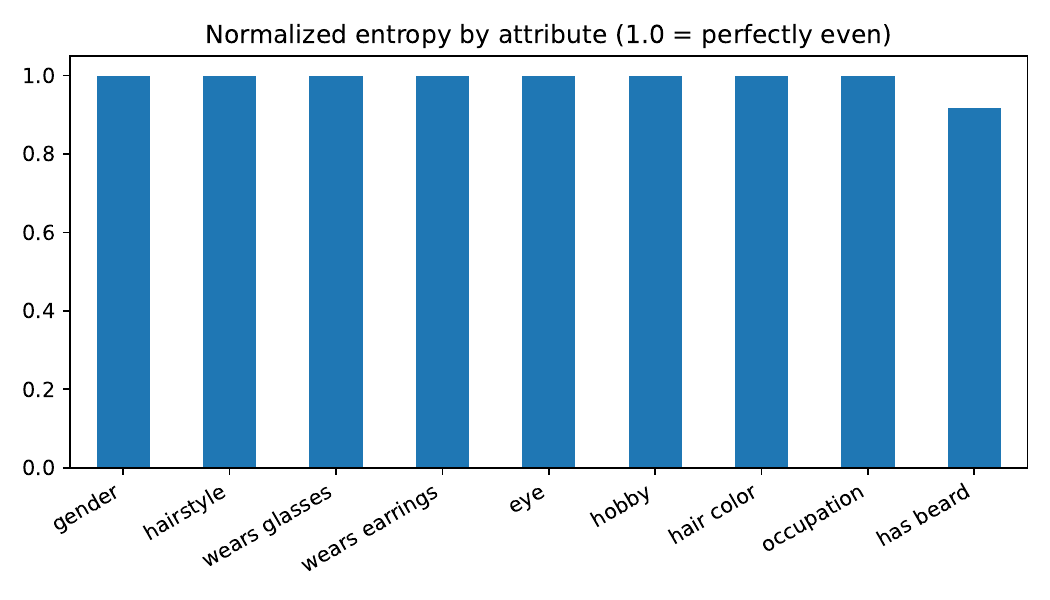}
   \caption{The Normalized Entropy of attributes.}
   \label{fig:guesswho_entropy}
\end{figure}

We design a set of 36 characters (C01–C36), with each character annotated by 9 discrete attributes shown in \cref{tab:attribute_universe}. The complete character set is detailed in \cref{tab:character_data}. To quantify the distributional balance of each attribute, we use normalized entropy (NE), where a value of 1.0 indicates a perfectly uniform distribution. As shown in \cref{fig:guesswho_entropy}, all attributes achieve $\mathrm{NE}=1.0$ except \texttt{has beard}. The lower $\mathrm{NE}$ for the \texttt{has beard} attribute is a result of our modeling constraint that only male characters can have beards. 

\begin{table}[t]
\centering
\caption{The Character Attribute Universe.}
\label{tab:attribute_universe}
\resizebox{\columnwidth}{!}{%
\begin{tabular}{lp{4cm}}
\toprule
\textbf{Attribute} & \textbf{Possible Values} \\
\midrule
\texttt{gender} & \texttt{male, female} \\
\texttt{hair color} & \texttt{red, blonde, black, white, brown} \\
\texttt{hairstyle} & \texttt{curly, short, long, bald} \\
\texttt{wears glasses} & \texttt{no, yes} \\
\texttt{has beard} & \texttt{no, yes} \\
\texttt{eye color} & \texttt{amber, brown, green, blue} \\
\texttt{hobby} & \texttt{movies, photography, music, games, reading, sports} \\
\texttt{wears earrings} & \texttt{no, yes} \\
\texttt{occupation} & \texttt{police, student, teacher, chef, doctor} \\
\bottomrule
\end{tabular}
}
\end{table}

\begin{table*}[t]
\centering
\caption{Guess Who character data.}
\setlength{\tabcolsep}{3pt}
\label{tab:character_data}
\begin{tabular}{llllllllll}
\toprule
\textbf{name} & \textbf{gender} & \textbf{hair color} & \textbf{hairstyle} & \textbf{wears glasses} & \textbf{has beard} & \textbf{eye} & \textbf{hobby} & \textbf{wears earrings} & \textbf{occupation} \\
\midrule
C01 & male & red & curly & no & no & amber & movies & no & police \\
C02 & male & blonde & short & no & no & brown & photography & no & student \\
C03 & male & black & long & no & yes & brown & music & yes & teacher \\
C04 & female & white & long & yes & no & amber & games & no & police \\
C05 & female & brown & short & no & no & green & reading & yes & student \\
C06 & female & brown & long & no & no & green & sports & no & police \\
C07 & female & black & curly & yes & no & blue & sports & no & chef \\
C08 & female & blonde & short & yes & no & green & music & yes & chef \\
C09 & female & black & long & yes & no & blue & reading & yes & teacher \\
C10 & female & white & short & no & no & amber & music & no & chef \\
C11 & female & black & short & yes & no & blue & games & no & chef \\
C12 & male & red & bald & no & no & blue & reading & no & chef \\
C13 & male & blonde & bald & yes & yes & amber & music & no & doctor \\
C14 & female & red & curly & yes & no & blue & photography & yes & student \\
C15 & male & blonde & bald & no & yes & brown & reading & yes & police \\
C16 & female & red & long & yes & no & brown & music & no & doctor \\
C17 & male & white & bald & yes & yes & blue & reading & no & teacher \\
C18 & female & white & long & no & no & green & games & yes & student \\
C19 & male & white & bald & yes & yes & green & games & yes & teacher \\
C20 & female & red & short & no & no & green & reading & yes & doctor \\
C21 & male & white & bald & yes & no & green & photography & yes & teacher \\
C22 & female & blonde & long & yes & no & amber & movies & no & student \\
C23 & male & black & short & no & no & brown & photography & yes & police \\
C24 & female & brown & curly & yes & no & blue & music & no & doctor \\
C25 & male & blonde & curly & no & yes & blue & movies & yes & police \\
C26 & female & red & short & no & no & green & sports & no & doctor \\
C27 & female & brown & curly & no & no & amber & movies & yes & chef \\
C28 & female & white & short & no & no & brown & games & yes & chef \\
C29 & male & brown & bald & yes & yes & amber & movies & yes & teacher \\
C30 & female & black & curly & no & no & amber & sports & no & teacher \\
C31 & male & black & bald & yes & no & brown & movies & no & police \\
C32 & male & white & long & yes & yes & blue & photography & yes & teacher \\
C33 & male & brown & curly & yes & yes & amber & photography & yes & student \\
C34 & male & brown & bald & yes & yes & brown & sports & no & student \\
C35 & male & blonde & long & no & yes & brown & games & yes & doctor \\
C36 & male & red & curly & no & yes & green & sports & no & doctor \\
\bottomrule
\end{tabular}
\end{table*}

\section{Implementation Details}

\subsection{SFT for Guess Number task}
\label{sec:sft_guessnum}
\mypar{Synthetic Dialogue Dataset Construction.}
We construct a synthetic dialogue dataset for the Guess Number task to support SFT. As detailed in \Cref{alg:guessnum}, the generation process for each sample begins by defining a random integer interval $\mathcal{C}$ whose length is sampled uniformly between 100 and 300. A hidden target number is then uniformly sampled from this interval. The current hypothesis space is represented by the candidate set $\mathcal{C}_k$. To manage task complexity, the agent's inquiries are restricted to six predefined question templates. These templates are $\operatorname{odd}$, $\operatorname{even}$, $<$, $\le$, $>$, and $\ge$. At each turn, the algorithm calculates EIG for each template over the current candidate set $\mathcal{C}_k$. The question with the highest EIG is subsequently selected. To improve diversity, if multiple questions score within a near-optimal range defined by a tolerance of $\tau = 10^{-4}$, one is sampled uniformly from that set. The agent makes a definitive guess once the candidate set size $\mathcal{C}_k$ is reduced to two or fewer. This procedure effectively implements a binary search policy over the hypothesis space.

For training purposes, each complete dialogue trajectory is expanded into multiple instances. Specifically, a dialogue of $n$ turns is segmented into $n$ separate training examples, where each example includes the full dialogue history up to that turn. The resulting corpus contains 16,804 training instances and 1,277 validation instances.

\begin{algorithm*}[h!]
\caption{Optimal Dialogue Synthesis for Guess Number task}
\label{alg:guessnum}
\begin{algorithmic}[1]
\Require $M$: number of dialogues to generate
\Require $T_{\max}$: maximum turns per dialogue (e.g., 16)
\Require $\tau$: tolerance for near-optimal question sampling (e.g., $0.0001$)
\Require $P$: six type of question templates ($\operatorname{odd}$, $\operatorname{even}$, $<$, $\leq$, $>$, $\geq$)
\Ensure $\mathcal{D}$: unfolded dialogue dataset
\State $\mathcal{D}\gets\emptyset$
\For{$m=1$ to $M$}
  \State $\text{start} \gets \operatorname{RamdomInt(100, 500)}$; \quad $N \gets \operatorname{RandomInt(100, 300)}$
  \State $\mathcal{C} = \gets \{\text{start}, \text{start} + 1, \dots, \text{start} + N - 1\}$; $c_{target} \leftarrow \text{Sample}(\mathcal{C})$ \Comment{Select a secret target}
  \State $\mathcal{C}_k \leftarrow \mathcal{C}$\Comment{Initialize candidate pool with all characters}
  \State $H\gets[\,]$ 
  \While{$|\mathcal{C}|>1$ \textbf{and} $|H|<T_{\max}$}
    \If{$|\mathcal{C}_k| <= 2$}
      \State $g\gets \text{RandomSample}(\mathcal{C}_k)$
      \If{$g$ = TRUE} 
        \State Append $(\text{GuessText}(g),\text{yes})$ to $H$;
        \textbf{break} 
       \Else 
          \State Append $(\text{GuessText}(g),\text{no})$ to $H$
          \State $\mathcal{C}_k \gets \{c \in \mathcal{C}_k \mid c \notin g\}$;
         \textbf{continue} 
        \EndIf
    \EndIf
    \State $\mathcal{Q}_{scores} \leftarrow \emptyset$ \Comment{Store scores for all potential questions this turn}
    \For{each question type $p \in P$} \Comment{Question scoring phase}
        \State $q \gets (p, \operatorname{Median}(\mathcal{C}_k))$ \Comment{Create a question using the median value of $\mathcal{C}_k$} 
        \State $\text{score} \leftarrow \operatorname{EIG}(\mathcal{C}_k, q)$ \Comment{Calculate EIG} 
        \State Add $(q, \text{score})$ to $\mathcal{Q}_{scores}$ 
    \EndFor
    \State $\mathcal{Q}_{\text{best}} \leftarrow \{q \mid (q, \text{score}) \in \mathcal{Q}_{scores} \text{ and } \text{score}_{\max} - \text{score} \le \tau\}$
    \State $q_{\text{chosen}} \leftarrow \text{RandomSample}(\mathcal{Q}_{\text{best}})$ \Comment{Select from near-optimal questions}
    \State $answer = \operatorname{Answer}(q, c_{target})$
    \State Append $(q, answer)$ to $H$; Update $\mathcal{C}_k$
  \EndWhile
  \State $\mathcal{D} \leftarrow \text{UnfoldTrajectories}(H)$
\EndFor
\State \textbf{return} $\mathcal{D}$
\end{algorithmic}
\end{algorithm*}

\mypar{Supervised Finetuning.}

We employ SFT on the Qwen2.5-7B-Instruct model with LoRA. We set the rank to 16 and alpha to 32.  The model is trained for 1 epoch using the AdamW optimizer. The optimization process incorporates a weight decay of 0.1 and a learning rate of $5 \times 10^{-6}$, managed by a cosine learning rate schedule. The training objective is a standard autoregressive loss, computed exclusively on the tokens generated by the model.

\begin{figure}[h!]
  \centering
   \includegraphics[page=1, width=0.9\linewidth]{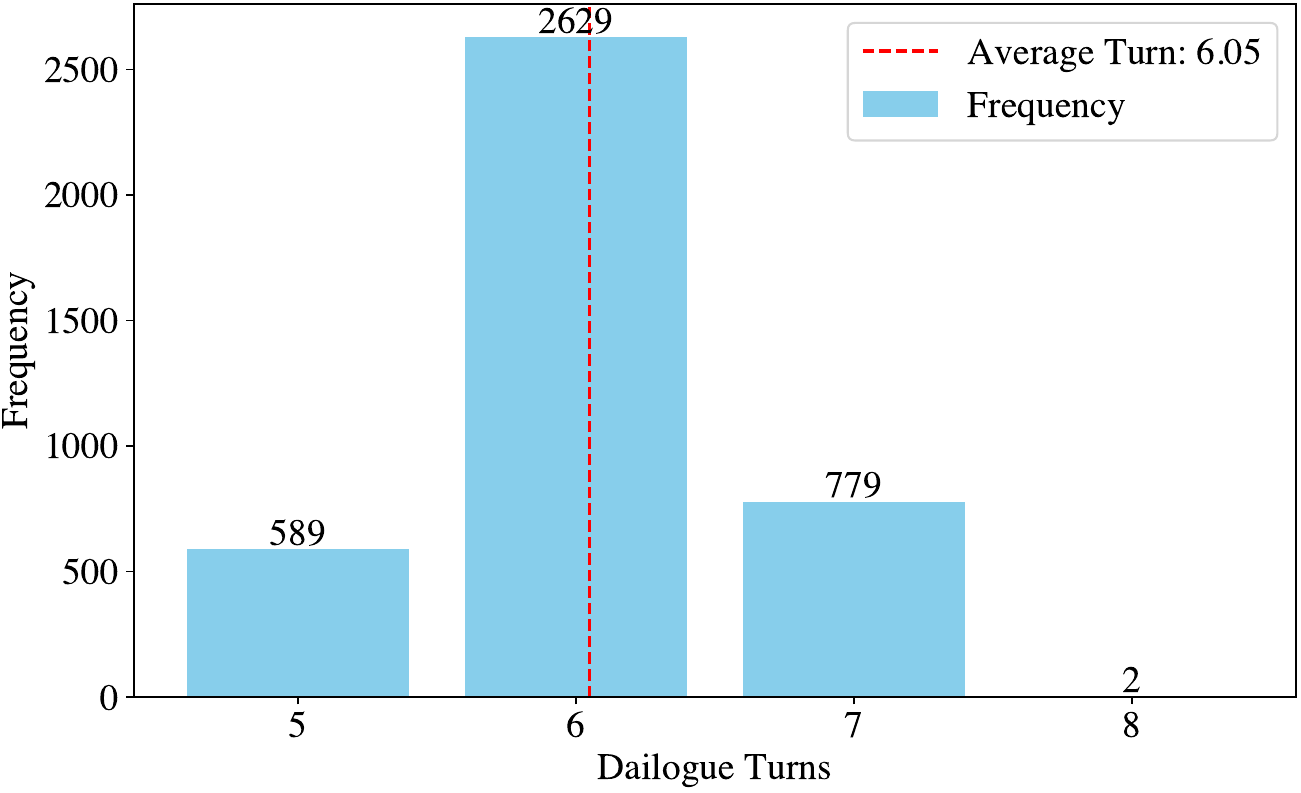}
   \caption{Distribution of Dialogue Turns.}
   \label{fig:dia_turn}
\end{figure}

\subsection{SFT for Guess Who task}
\label{sec:sft_guesswho}

\begin{algorithm*}[h!]
\caption{Optimal Dialogue Synthesis for Guess Who task}
\label{alg:guesswho_synthesis}
\begin{algorithmic}[1]
\Require
    $M$: Number of dialogues to generate
\Require
    $K$: Candidate threshold to start guessing (e.g., 2)
\Require
    $\mathcal{A}$: Set of all possible attributes (e.g., hair color, gender)
\Require
    $\mathcal{V}_a$: Set of possible values for attribute $a \in \mathcal{A}$
\Require
    $\mathcal{P}$: A pre-computed mapping of canonical questions to paraphrases
\Require
    $\tau$: Score tolerance for near-optimal question sampling (e.g., 0.0001)
\Require    
    $T_{\max}$: Maximum turns per dialogue (e.g., 16)
\Ensure
    $\mathcal{D}$: A dataset of unfolded dialogue trajectories

\State $\mathcal{D} \leftarrow \emptyset$ \Comment{Initialize the final dataset}
\For{$m = 1$ to $M$} \Comment{Loop to generate M full dialogues}
    \State $\mathcal{C} \leftarrow \text{GenerateRandomCharacterSet}(\mathcal{A}, \{\mathcal{V}_a\})$ \Comment{Create a new game board}
    \State $c_{target} \leftarrow \text{RandomSample}(\mathcal{C})$ \Comment{Select a secret target character}
    \State $\mathcal{C}_k \leftarrow \mathcal{C}$ \Comment{Initialize candidate pool with all characters}
    \State $H \leftarrow []$ \Comment{Initialize empty dialogue history for this game}

    \While{$|\mathcal{C}_k| > 1$ \textbf{ and } $|H| < T_{\max}$}
        \If{$|\mathcal{C}_k| \le K$} \Comment{Guessing phase}
            \State $c_{guess} \leftarrow \text{RandomSample}(\mathcal{C}_k)$
            \State Update $H$ with final guess and outcome.
            \State \textbf{break} \Comment{End the current game}
        \EndIf
        
        \State $\mathcal{Q}_{scores} \leftarrow \emptyset$ \Comment{Store scores for all potential questions this turn}
        
        \For{each attribute $a \in \mathcal{A}$} \Comment{Question scoring phase} 
            \For{each value $v \in \mathcal{V}_a$} 
                \State $q \leftarrow (a, v)$ 
                \If{$q$ was not asked before in $H$} 
                    \State $\text{score} \leftarrow \operatorname{EIG}(\mathcal{C}_k, q)$ \Comment{Calculate EIG} 
                    \State Add $(q, \text{score})$ to $\mathcal{Q}_{scores}$ 
                \EndIf 
            \EndFor
        \EndFor
        \State $\mathcal{Q}_{\text{best}} \leftarrow \{q \mid (q, \text{score}) \in \mathcal{Q}_{scores} \text{ and } \text{score}_{\max} - \text{score} \le \tau\}$
        \State $q_{\text{chosen}} \leftarrow \text{RandomSample}(\mathcal{Q}_{\text{best}})$ \Comment{Select from near-optimal questions}

        \State $q_{\text{paraphrased}} \leftarrow \text{GPT}(q_{\text{chosen}})$ \Comment{Linguistic variation}
        
        \State $answer \leftarrow (c_{target}[q_{\text{chosen}}.a] == q_{\text{chosen}}.v)$ \Comment{Get true answer from target}
        \State Append $(q_{\text{paraphrased}}, answer)$ to $H$
        
        \If{$answer$ is TRUE} \Comment{Update candidate pool}
            \State $\mathcal{C}_k \leftarrow \{c \in \mathcal{C}_k \mid c[q_{\text{chosen}}.a] = q_{\text{chosen}}.v\}$
        \Else
            \State $\mathcal{C}_k \leftarrow \{c \in \mathcal{C}_k \mid c[q_{\text{chosen}}.a] \neq q_{\text{chosen}}.v\}$
        \EndIf
    \EndWhile
    \State $\mathcal{D} \leftarrow \text{UnfoldTrajectories}(H)$
\EndFor
\State \textbf{return} $\mathcal{D}$
\end{algorithmic}
\end{algorithm*}

\mypar{Synthetic Dialogue Dataset Construction.}
We construct a synthetic dialogue dataset for Guess Who task. Following the procedure outlined in \Cref{alg:guesswho_synthesis}, each dialogue commences by sampling a board with 36 characters, each possessing a set of discrete attributes. A single character is then selected as the hidden target. The system maintains a candidate set $\mathcal{C}_k$ which is progressively reduced after each turn. At every step, the system scores all unasked single-value questions and selects one to pose.

Our approach employs a greedy policy based on EIG. For any single-value question $q=(a, v)$, we compute its $\mathrm{EIG}(q)$ from the number of candidates $\mathcal C_k$. The question with the highest EIG score is subsequently chosen. To introduce strategic diversity, if multiple questions fall within a near-optimal band defined by a tolerance $\tau$, which is set to 0.01\% in our experiments, we sample uniformly among them. The system provides a binary response based on the ground-truth target and prunes the candidate set accordingly. The dialogue stops and makes a final guess when the number of remaining candidates is at most $K$ (e.g., $K=2$). The distribution of dialogue lengths is presented in \Cref{fig:dia_turn}, showing a mean of 6.05 turns. This policy locally maximizes information gain to rapidly compress the hypothesis space, while mild randomization prevents policy degeneracy.

To enhance the scale and linguistic diversity of the dataset, we generate dialogues by repeatedly resampling boards and targets. We then use GPT-4o to generate semantically equivalent paraphrases for each question. Same as the guess number task, each full dialogue trajectory is expanded into multiple training samples. The final corpus consists of 24,105 training instances and 1,034 validation instances.

\mypar{Supervised Finetuning.}
The SFT training procedure and hyperparameters are identical to those used for the Guess Number task (\Cref{sec:sft_guessnum}).

\subsection{Fine-tuning BLIP}
\label{sec:blip}
\mypar{Synthetic Dialogue Dataset Construction.}
To address the absence of a public corpus for our task, we employ a programmatic synthesis approach to generate a custom-built dataset. The CelebA dataset served as an ideal foundation, as its binary attributes are directly applicable for generating the yes/no dialogues.

Our synthesis pipeline begins by creating a question-answer (QA) pair for each of the 40 attributes associated with an image. To enhance linguistic diversity, we manually prepare three distinct phrasings for each attribute's corresponding question. During generation, one of these three templates is selected at random. A positive QA pair is generated if the attribute value is $1$, while a negative QA pair is generated if the value is $-1$.

\begin{figure}[t]
  \centering
   \includegraphics[page=7, width=1\linewidth]{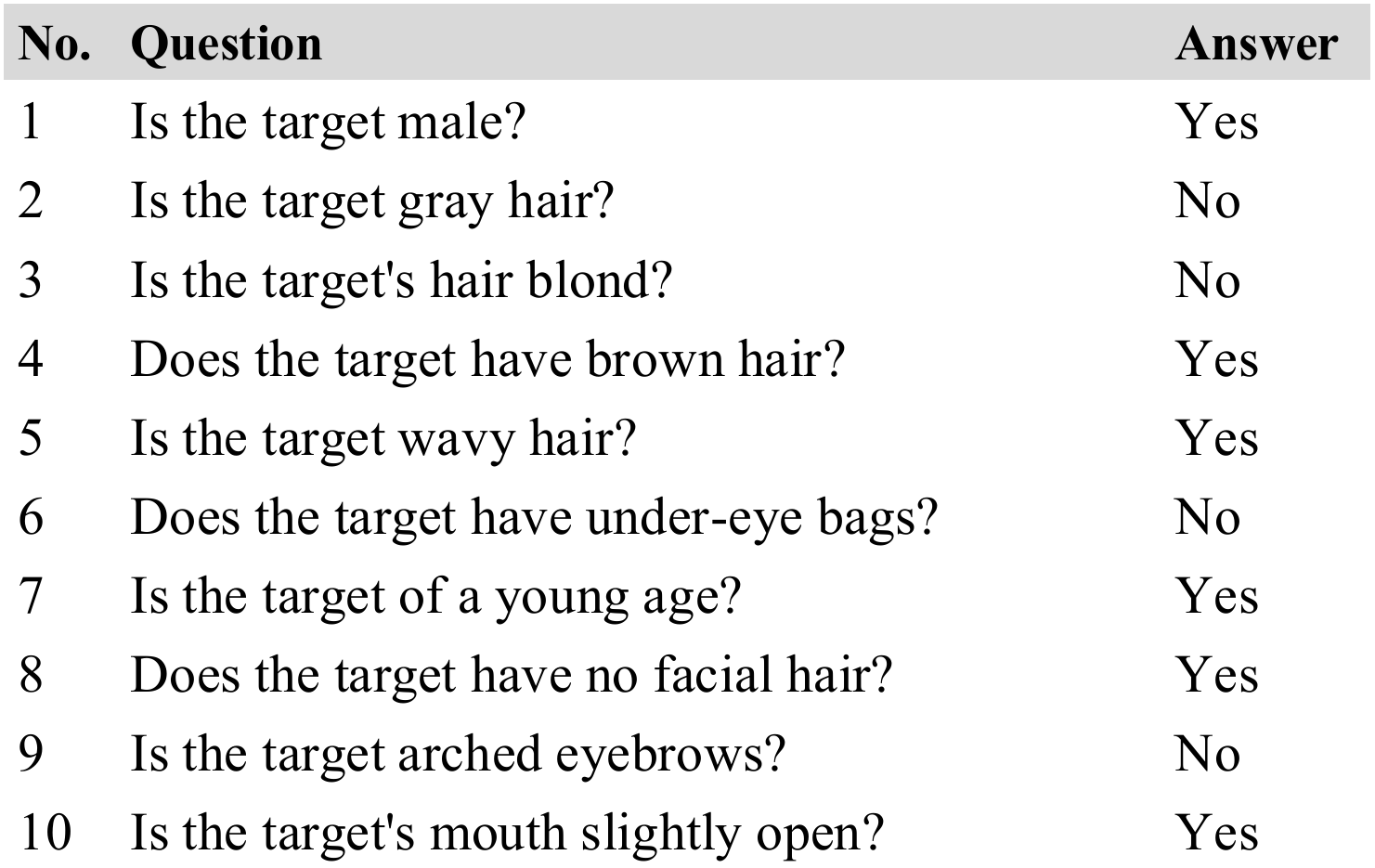}
   \caption{An example of synthetic dialogue.}
   \label{fig:synthetic_dialogue}
\end{figure}

To construct a full dialogue, we first select images with no more than 20 positive attributes. The corresponding positive QA pairs are then randomly shuffled to form the dialogue's core. Rather than padding all dialogues to a fixed length, we use a stochastic process to determine the number of negative questions. For an image $I_i$ with $m$ positive attributes, we sample an integer $x$ from the interval $[0,20-m]$ and then randomly select $x$ negative QA pairs. Subsequently, we apply a negative sampling strategy to insert these negative QA pairs among the positive ones. This approach preferentially places a related negative question before a positive one (e.g., asking about ``Blond Hair" or ``Gray Hair" before ``Black Hair"). This method is designed to simulate a realistic process of elimination, where the scope of inquiry is progressively narrowed to identify the target attributes. In total, we generate 40,000 dialogues using this procedure. \cref{fig:synthetic_dialogue} shows an example of a generated dialogue.

\mypar{Supervised Fine-tuning.}
We use the official implementation \footnote{BLIP codebase \url{https://github.com/salesforce/BLIP}} to fine-tune the BLIP base model. The model is pre-trained on the Flickr30k dataset \cite{young-etal-2014-image}. We use the AdamW optimizer with an initial learning rate of $5\times10^{-6}$, a weight decay of 0.05, and a training batch size of 32. While the model is fine-tuned for a total of 12 epochs, the checkpoint from the 9th epoch is selected for the final evaluation, as it achieves optimal retrieval performance on the validation set.
We implement data augmentation to improve the model's generalization and robustness. For image inputs, we apply standard transformations, including random horizontal flipping, adjustments to brightness and sharpness, random translation and rotation, and normalization. On the text side, we apply an augmentation method termed "Random Round Retention". This method randomly truncates dialogues, retaining only the first $N > 0$ QA rounds as input. The objective is to enhance the model's adaptability to conversations of varying lengths. Additionally, since our dialogue texts significantly exceed the default 50-character limit in the official BLIP implementation, we expand the maximum input text length to 300. This modification ensures that the complete conversational context is processed by the model.

\end{document}